\documentclass{article}

\pdfoutput=1

\usepackage[nonatbib,final]{neurips_2024}

\usepackage[utf8]{inputenc} 
\usepackage[T1]{fontenc}    
\usepackage{url}            
\usepackage{booktabs}       
\usepackage{amsfonts}       
\usepackage{nicefrac}       
\usepackage{microtype}      

\usepackage{xcolor}

\usepackage{amsmath}
\usepackage{bbm}
\usepackage{multirow}
\usepackage{bbding}

\usepackage[pdftex]{graphicx}
\usepackage{float}
\usepackage{subfig}
\usepackage{amssymb}

\usepackage{wrapfig}

\usepackage[colorlinks,linkcolor=red,filecolor=blue,citecolor=blue,urlcolor=blue]{hyperref}

\newcommand{\etal}{{\emph{et al.}}}
\DeclareMathOperator*{\argmin}{argmin}

\title{Continual Audio-Visual Sound Separation}

\author{Weiguo Pian\textsuperscript{\rm 1} \, Yiyang Nan\textsuperscript{\rm 2} \, Shijian Deng\textsuperscript{\rm 1} \, Shentong Mo\textsuperscript{\rm 3} \, Yunhui Guo\textsuperscript{\rm 1} \, Yapeng Tian\textsuperscript{\rm 1} \\
\\
\textsuperscript{\rm 1} The University of Texas at Dallas \, \textsuperscript{\rm 2} Brown University \, \textsuperscript{\rm 3} Carnegie Mellon University \\
}

\begin{document}

\maketitle

\begin{abstract}

In this paper, we introduce a novel continual audio-visual sound separation task, aiming to continuously separate sound sources for new classes while preserving performance on previously learned classes, with the aid of visual guidance. This problem is crucial for practical visually guided auditory perception as it can significantly enhance the adaptability and robustness of audio-visual sound separation models, making them more applicable for real-world scenarios where encountering new sound sources is commonplace. The task is inherently challenging as our models must not only effectively utilize information from both modalities in current tasks but also preserve their cross-modal association in old tasks to mitigate catastrophic forgetting during audio-visual continual learning. To address these challenges, we propose a novel approach named ContAV-Sep (\textbf{Cont}inual \textbf{A}udio-\textbf{V}isual Sound \textbf{Sep}aration). ContAV-Sep presents a novel Cross-modal Similarity Distillation Constraint (CrossSDC) to uphold the cross-modal semantic similarity through incremental tasks and retain previously acquired knowledge of semantic similarity in old models, mitigating the risk of catastrophic forgetting. The CrossSDC can seamlessly integrate into the training process of different audio-visual sound separation frameworks. Experiments demonstrate that ContAV-Sep can effectively mitigate catastrophic forgetting and achieve significantly better performance compared to other continual learning baselines for audio-visual sound separation. Code is available at: \url{https://github.com/weiguoPian/ContAV-Sep_NeurIPS2024}.

\end{abstract}
    
\section{Introduction}


Humans can effortlessly separate and identify individual sound sources in daily experience~\cite{haykin2005cocktail,bregnian1993auditory,sussman2005integration,kushwaha2022analyzing}. This skill plays a crucial role in our ability to understand and interact with the complex auditory environments that surround us~\cite{kushwaha2023multimodal}. However, replicating this capability in machines remains a significant challenge due to the inherent complexity of real-world auditory scenes~\cite{bregnian1993auditory,weisser2018complex}. Inspired by the multisensory perception of humans~\cite{stein2008multisensory,spence2007audiovisual}, audio-visual sound separation tackles this challenge by utilizing visual information to guide the separation of individual sound sources in an audio mixture.

Recent advances in deep learning have led to significant progress in audio-visual sound separation~\cite{zhao2018sound,gao2019co,gan2020music,tian2021cyclic,chen2023iquery,tan2023language,ye2023lavss,su2023separating,chatterjee2022learning,tzinis2022audioscopev2}.
Benefiting from more advanced architectures (\textit{e.g.,} U-Net~\cite{zhao2018sound,gao2019co}, Transformer~\cite{chen2023iquery}, and diffusion models~\cite{huang2023davis}) and discriminative visual cues (\textit{e.g.,} grounded visual objects~\cite{tian2021cyclic}, motion~\cite{zhao2019sound}, and dynamic gestures~\cite{gan2020music}), audio-visual separation models are able to separate sounds ranging from domain-specific speech, musical instrument sounds to open-domain general sounds within training sound categories.
However, a limitation of these studies is their focus on scenarios where all sound source classes are presently known, overlooking the potential inclusion of unknown sound source classes during inference in real-world applications. This oversight leads to the \textit{catastrophic forgetting} issue~\cite{kirkpatrick2017overcoming, aljundi2018memory}, where the fine-tuning of models on new classes detrimentally impacts their performance on previously learned classes. Despite Chen et al.~\cite{chen2023iquery} demonstrating that their iQuery model can generalize to new classes well through simple fine-tuning, it still suffers from the catastrophic forgetting problem on old classes. This prevents the trained models from continuously updating in real-world scenarios, impeding their adaptability to dynamic environments.
The question \textit{how to effectively leverage visual guidance to continuously separate sounds from new categories while preserving separation ability for old sound categories} remains open.

\begin{wrapfigure}[24]{r}{0.59\textwidth}
  \centering
  \vspace{-3mm}
   \includegraphics[width=0.55\textwidth]{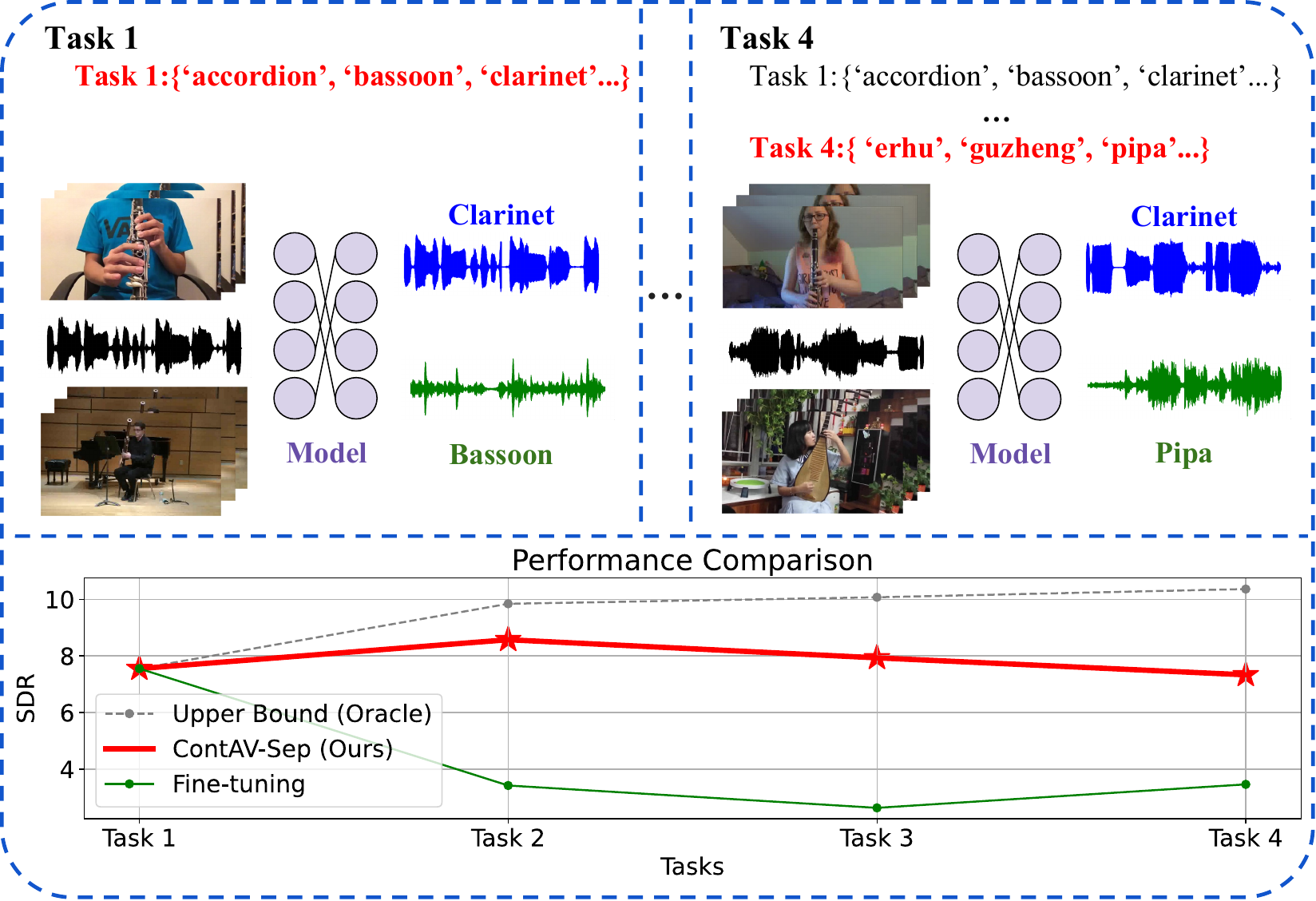}
  \caption{\textbf{Top}: Illustration of the continual audio-visual sound separation task, where the model (separator) learns from sequential audio-visual sound separation tasks. \textbf{Bottom}: Illustration of the catastrophic forgetting problem in continual audio-visual sound separation and its mitigation by our proposed method. Fine-tuning: Directly fine-tune the separation model on new sound source classes; Upper bound: Train the model using all training data from seen sound source classes.}
  \label{fig:teaser}
\end{wrapfigure}


To bridge this gap, we introduce a novel \textit{continual audio-visual sound separation} task by integrating audio-visual sound separation with continual learning principles. The goal of this task is to develop an audio-visual model that can continuously separate sound sources in new classes while maintaining performance on previously learned classes. The key challenge we need to address is catastrophic forgetting during continual audio-visual learning, which occurs when the model is updated solely with data from new classes or tasks, resulting in a significant performance drop on old ones. We illustrate our new task and the catastrophic forgetting issue in Fig.~\ref{fig:teaser}.

Unlike typical continual learning problems such as task-, domain-, or class-incremental classification in visual domains~\cite{ahn2021ssil,rebuffi2017icarl,li2017learning,pian2023audio,zhou2023deep}, which result in progressively increasing logits (or probability distribution) across all observed classes at each incremental step, our task uniquely produces fixed-size separation masks throughout all incremental steps. In this context, each entry in the mask does not directly correspond to any specific classes. Additionally, the new task involves both audio and visual modalities. Therefore, simply applying existing visual-only methods cannot fully exploit and preserve the inherent cross-modal semantic correlations. Very recently, Pian \etal~\cite{pian2023audio} and Mo \etal~\cite{mo2023class} extended continual learning to the audio-visual domain, but both focused on classification tasks. 

To address these challenges, in this paper, we propose a novel approach named ContAV-Sep (\textbf{Cont}inual \textbf{A}udio-\textbf{V}isual Sound \textbf{Sep}aration). Upon the framework, we introduce a novel \textit{Cross-modal Similarity Distillation Constraint (CrossSDC)} to not only maintain the cross-modal semantic similarity through incremental tasks but also preserve previously learned knowledge of semantic similarity in old models to counter catastrophic forgetting. The CrossSDC is a generic constraint that can be seamlessly integrated into the training process of different audio-visual sound separators.
To evaluate the effectiveness of our proposed ContAV-Sep, we conducted experiments on the MUSIC-21 dataset within the framework of continual learning, using the state-of-the-art audio-visual sound separation model iQuery~\cite{chen2023iquery} and a representative audio-visual sound separation model Co-Separation~\cite{gao2019co}, as our separation base models. Experiments demonstrate that ContAV-Sep can effectively mitigate catastrophic forgetting and achieve significantly better performance than other continual learning baselines. In summary, this paper contributes follows:

\textbf{(i)} To explore more practical audio-visual sound separation, in which the separation model should be generalized to new sound source classes continually, we pose a \textit{Continual Audio-Visual Sound Separation} task that trains the separation model under the setting of continual learning. To the best of our knowledge, this is the first work on continual learning for audio-visual sound separation.

\textbf{(ii)} We propose ContAV-Sep for the new task. It uses a novel cross-modal similarity distillation constraint to preserve cross-modal semantic similarity knowledge from previously learned models.

\textbf{(iii)} Experiments on the MUSIC-21 dataset can validate the effectiveness of our ContAV-Sep, demonstrating promising performance gain over baselines.

\section{Related Work}
\textbf{Audio-Visual Sound Separation.}
Audio-visual sound separation aims to separate individual sound sources from an audio mixture guided by visual cues. A line of research emerges under various scenarios, such as separating musical instruments \cite{gao2019co,zhao2018sound,xu2019recursive, gan2020music,zhao2019sound,tian2021cyclic}, human speech \cite{gabbay2017visual,afouras2018conversation,ephrat2018looking,owens2018audio,chung2020facefilter}, or sound sources in in-the-wild videos \cite{gao2018learning,tzinis2020into}. Many frameworks and methods have been proposed to address challenges within specific problem settings. For instance, the extraction of face embeddings proves beneficial for speech audio separation \cite{ephrat2018looking}. Moreover, incorporating object detection can provide an additional advantage \cite{gao2019co,gao2018learning}. The utilization of trajectory optical flows to leverage temporal motion information in videos, as demonstrated by \cite{zhao2019sound}, also yields improvements. In this work, not competing on designing stronger separators, we would advance the exploration of the audio-visual sound separation within the paradigm of continual learning. We investigate how a model can learn to consistently separate sound sources from sequential separation tasks without forgetting previously acquired knowledge.

\textbf{Continual Learning.}
The field of continual learning has drawn significant attention, especially in visual domains, with various approaches addressing this challenge. Notable among these are regularization-based methods, exemplified in works such as \cite{kirkpatrick2017overcoming, aljundi2018memory,kim2023achieving,liang2023adaptive}. These approaches involve applying regularization to crucial parameters associated with old tasks to maintain the model's capabilities and during incremental steps, less important parameters are given higher priority for updates compared to important ones. 
Conversely, several works~\cite{rebuffi2017icarl,castro2018end,belouadah2019il2m,hou2019learning,chaudhry2019tiny,prabhu2020gdumb,buzzega2020dark,cha2023rebalancing,luo2023class} applied rehearsal-based pipelines to enable the model review previously learned knowledge. 
For instance, Rebuffi~\textit{et al.}~\cite{rebuffi2017icarl} proposed one of the most representative exemplars selection strategy Nearest-Mean-of-Exemplars (NME), selects the most representative exemplars in each class based on the distance to the feature center of the class. Meanwhile, pseudo-rehearsal~\cite{odena2017conditional, ostapenko2019learning,tang2022learning} employs generative models to create pseudo-exemplars based on the estimated distribution of data from previous classes. Moreover, architecture-based/dynamic architecture methods\cite{pham2021dualnet,arani2022learning,nie2023bilateral,golkar2019continual, hung2019compacting,li2019learn,luo2023class} proposed to modify the model architecture itself to enable the model to acquire new knowledge while mitigating the forgetting of old knowledge. Specifically, Pham~\textit{et al.}~\cite{pham2021dualnet} proposed a dual network architecture, in which one is to learn new tasks while the other one is for retaining knowledge learned from old tasks. Wang~\textit{et al.}~\cite{wang2022foster} combined the dynamic architecture and distillation constraint to mitigate the issue of continual-increasing overhead problem in dynamic architecture-based continual learning method. 
However, above studies mainly concentrate on the area of continual image classification. 
Recently, researchers also explored other continual learning scenarios beyond image classification. For instance, Park~\textit{et al.}~\cite{park2021class} extend the knowledge distillation-based~\cite{ahn2021ssil,douillard2020podnet} continual image classification method to the domain of video by proposing the time-channel distillation constraint. Douillard~\textit{et al.}~\cite{douillard2021plop} proposed to tackle the continual semantic segmentation task with multi-view feature distillation and pseudo-labeling. Xiao~\textit{et al.}~\cite{xiao2023endpoints} further addressed the continual semantic segmentation problem through weights fusion strategy between old and current models. Wang~\textit{et al.}~\cite{wang2019continual} addressed the continual sound classification task through generative replay.
Furthermore, continual learning has also been explored in the domain of language/vision-language learning tasks~\cite{ke2023continual,mi2020continual,scialom2022fine,srinivasan2022climb,fini2022self,zhou2023learning}, self-supervised representation learning~\cite{fini2022self, madaan2022representational,purushwalkam2022challenges,yan2022generative,lee2023lifelong,wang2022learningrepresentations}, audio classification~\cite{bhatt2024characterizing,wang2021few} and fake audio detection~\cite{ma2021continual,zhang2023you}, etc. Despite the success of existing continual learning methods in various scenarios, their applicability in the domain of continual audio-visual sound separation is still unexplored. Although Pian~\textit{et al.}~\cite{pian2023audio} and Mo~\textit{et al.}~\cite{mo2023class} proposed to tackle the catastrophic problem in audio-visual learning, their studies mainly concentrated in the area of audio-visual video classification. In contrast to existing works in continual learning, in this paper, we delves into the continual audio-visual sound separation, aiming to tackle the challenge of catastrophic forgetting specifically in the context of separation mask prediction for complicated mixed audio signals within joint audio-visual modeling.

\section{Method}


\vspace{-2mm}
\subsection{Problem Formulation}
\label{subsec:problem}

\vspace{-2mm}
\textbf{Audio-Visual Sound Separation.}
Audio-visual sound separation aims to separate distinctive sound signals according to the given associated visual guidance. Following previous works~\cite{chen2023iquery,gao2019co,tian2021cyclic,gan2020music,xu2019recursive}, we adopt the common ``mix-and-separation'' training strategy to train the model. Given two videos $\boldsymbol{V}_1 (\boldsymbol{s}_1,\boldsymbol{v}_1)$ and $\boldsymbol{V}_2 (\boldsymbol{s}_2,\boldsymbol{v}_2)$, we can obtain the input mixed sound signal $\boldsymbol{S}$ by mixing two video sound signals $\boldsymbol{s}_1$ and $\boldsymbol{s}_2$, and then we can have the ratio masks $\boldsymbol{M}^1=\boldsymbol{s}_1/\boldsymbol{S}$ and $\boldsymbol{M}^2=\boldsymbol{s}_2/\boldsymbol{S}$\footnote{In practice, the audio signal is first processed using the Short-Time Fourier Transform (STFT) to generate a spectrogram. For brevity, we will denote spectrogram magnitudes as $\boldsymbol{s}_1$, $\boldsymbol{s}_2$, and $\boldsymbol{S}$.}. The goal of the task is to utilize the corresponding visual guidance $\boldsymbol{v}_1$ and $\boldsymbol{v}_2$ to predict the ratio masks for reconstructing the two individual audio signals.
This process can be formulated as:
\begin{equation}
    \begin{split}
        &\boldsymbol{\hat{M}}^1 = \mathcal{F}_{\boldsymbol{\Theta}}(\boldsymbol{S}, \boldsymbol{v}_1), \\
        &\boldsymbol{\hat{M}}^2 = \mathcal{F}_{\boldsymbol{\Theta}}(\boldsymbol{S}, \boldsymbol{v}_2),
    \end{split}
    \label{eq:separation}
\end{equation}
where $\mathcal{F}_{\boldsymbol{\Theta}}$ is the separation model with trainable parameters $\boldsymbol{\Theta}$. And then, the original sound signals $\boldsymbol{s}_1$ and $\boldsymbol{s}_2$ are used to calculate the loss function for optimizing the model:
\begin{equation}
        \boldsymbol{\Theta}^* = \argmin_{\boldsymbol{\Theta}} \mathbb{E}_{(\boldsymbol{V}_1,\boldsymbol{V}_2)\sim\mathcal{D}}\Big[\mathcal{L}(\boldsymbol{\hat{M}}^1, \boldsymbol{M}^1)+\mathcal{L}(\boldsymbol{\hat{M}}^2, \boldsymbol{M}^2)\Big],
    \label{eq:separation_loss}
\end{equation}
where $\mathcal{D}$ denotes the training set, and $\mathcal{L}$ is the loss function between the prediction and ground-truth.

\textbf{Continual Audio-Visual Sound Separation.} 
Our proposed continual audio-visual sound separation task aims to train a model $\mathcal{F}_{\boldsymbol{\Theta}}$ continually on a sequence of $T$ separation tasks $\{\mathcal{T}_1,\mathcal{T}_2,...,\mathcal{T}_T\}$. For the $t$-th task $\mathcal{T}_t$ (incremental step $t$), we have a training set $\mathcal{D}_t=\{ \boldsymbol{V}^i (\boldsymbol{s}^i, \boldsymbol{v}^i), y_t^i \}_{i=1}^{n_t}$, where $i$ and $n_t$ denote the $i$-th video sample and the total number of samples in $\mathcal{D}_t$ respectively, and $y_t^i\in \mathcal{C}_t$ is the corresponding sound source class of video $\boldsymbol{V}^i$, where $\mathcal{C}_t$ is the training sound class label space of task $\mathcal{T}_t$. For any two tasks $\mathcal{T}_{t_1}$ and $\mathcal{T}_{t_2}$ and their corresponding training sound class label space $\mathcal{C}_{t_1}$ and $\mathcal{C}_{t_2}$, we have $\mathcal{C}_{t_1}\cap\mathcal{C}_{t_2}=\emptyset$. Following previous works in continual learning~\cite{rebuffi2017icarl,ahn2021ssil,pian2023audio,mo2023class,kang2022class,wang2022learning}, for a task $\mathcal{T}_t$, where $t>1$, holding a small size of memory/exemplar set $\mathcal{M}_t$ to store some data from old tasks is permitted in our setting. Therefore, with the memory/exemplar set $\mathcal{M}_t$, all available data that can be used for training in task $\mathcal{T}_t$ ($t>1$) can be denoted as $\mathcal{D}'_t = \mathcal{D}_t\cup\mathcal{M}_t$. Finally, the training process of Eq.~\ref{eq:separation_loss} in our continual audio-visual sound separation setting can be denoted as:
\begin{equation}
    \begin{split}
        \boldsymbol{\Theta}_t = \argmin_{\boldsymbol{\Theta}_{t-1}} \mathbb{E}_{(\boldsymbol{V}_1,\boldsymbol{V}_2)\sim\mathcal{D}'_t}\Big[\mathcal{L}&(\boldsymbol{\hat{M}}^1, \boldsymbol{M}^1) + \mathcal{L}(\boldsymbol{\hat{M}}^2, \boldsymbol{M}^2)\Big], \\
        \textit{s.t.} \ \ \boldsymbol{\hat{M}}^1 =\mathcal{F}_{\boldsymbol{\Theta}_{t-1}}(\boldsymbol{S}, \boldsymbol{v}_1&),\ \boldsymbol{\hat{M}}^2 = \mathcal{F}_{\boldsymbol{\Theta}_{t-1}}(\boldsymbol{S}, \boldsymbol{v}_2),
    \end{split}
    \label{eq:cont_separation_loss}
\end{equation}
which means that the new model $\boldsymbol{\Theta}_t$ is obtained by updating the old model $\boldsymbol{\Theta}_{t-1}$ which was trained on the previous task, using the current task's available data $\mathcal{D}'_t$.
After the training process for task $\mathcal{T}_t$ with $\mathcal{D}'_t$, the updated model will be evaluated on a testing set which includes video samples from all seen sound source classes up to continual step $t$ (task $\mathcal{T}_t$). And the evaluation also follows the common “mix-and-separation” strategy. 
During this continual learning process, the model's separation performance on the previously learned tasks drops significantly after training on new tasks. This learning issue is referred to as the \textit{catastrophic forgetting}~\cite{kirkpatrick2017overcoming,li2017learning, aljundi2018memory} problem, which poses a considerable challenge in continual audio-visual sound separation. 

\subsection{Overview}
\label{subsec:overview}
\begin{figure*}[t]
    \centering
    \includegraphics[width=0.98\textwidth]{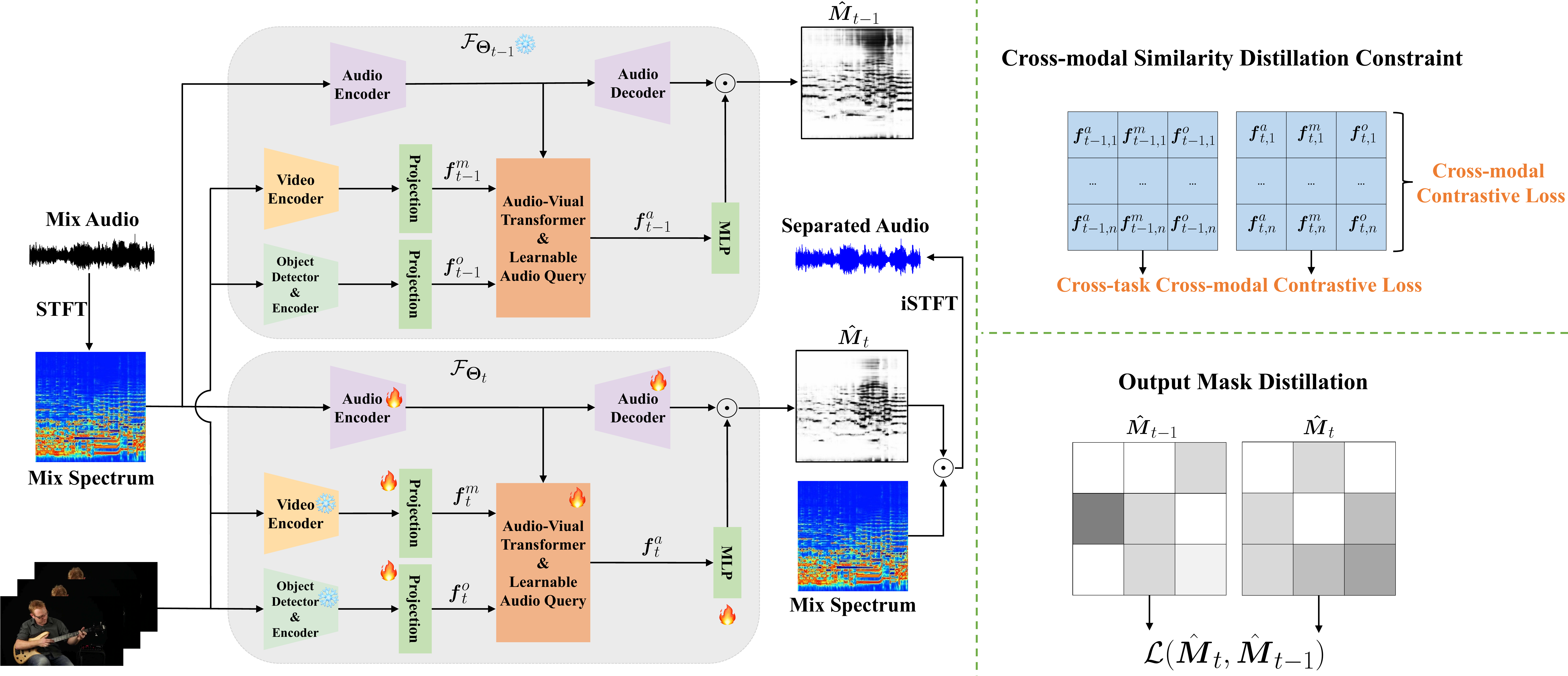}
    \caption{Overview of our proposed ContAV-Sep, which consists of an audio-visual sound separation base model architecture, an Output Mask Distillation, and our proposed Cross-modal Similarity Distillation Constraint. The fire icon denotes the module is trainable, while the snowflake icon denotes that the module is frozen. The (i)STFT stands for (inverse) Short-Time Fourier Transform. Please note that, the old model $\mathcal{F}_{\boldsymbol{\Theta}_{t-1}}$ is frozen during training.}
    \label{fig:overview}
\end{figure*}

To address the challenge of catastrophic forgetting in continual audio-visual sound separation, we introduce \textbf{ContAV-Sep}.
This new framework, illustrated in Fig.~\ref{fig:overview}, consists of three key components: a separation base model, an output mask distillation module, and our proposed \textit{Cross-modal Similarity Distillation Constraint (CrossSDC)}. We use a recent state-of-the-art audio-visual separator: iQuery~\cite{chen2023iquery} as the base model of our approach, which contains a video encoder to extract the global motion feature, an object detector and image encoder to obtain the object feature, a U-Net~\cite{ronneberger2015u} for mixture sound encoding and separated sound decoding, and an audio-visual Transformer to get the separated sound feature through multi-modal cross-attention mechanism and class-aware audio queries. For the object detector, we follow iQuery~\cite{chen2023iquery} and use the pre-trained Detic~\cite{zhou2022detecting}, a universal object detector, to detect the sound source objects in each frame. For the video encoder and the image encoder, inspired by the excellent generalization ability of recent self-supervised pre-trained models, which has been proven to be effective and appropriate in continual learning~\cite{pian2023audio}, we apply two self-supervised pre-trained models VideoMAE~\cite{tong2022videomae} and CLIP~\cite{radford2021learning} as the video encoder and the image encoder, respectively. Note that, during the training process, the object detector, video encoder, and image encoder are frozen.

Given a pair of videos $\boldsymbol{V}_1 (\boldsymbol{s}_1,\boldsymbol{v}_1)$ and  $\boldsymbol{V}_2 (\boldsymbol{s}_2,\boldsymbol{v}_2)$, at incremental step $t$ (task $\mathcal{T}_t$), the U-Net audio encoder $\mathcal{F}^{AE}_t$ takes the mixed audio signal $\boldsymbol{S}$ obtained by mixing $\boldsymbol{s}_1$ and $\boldsymbol{s}_2$ as input, and generates the latent mixed audio feature. This process can be expressed as:
\begin{equation}
    \begin{split}
        \boldsymbol{f}_t^{lat.} = \mathcal{F}^{AE}_t(\boldsymbol{S}),
    \end{split}
    \label{eq:unet_encoder}
\end{equation}
Then, the audio-visual Transformer $\mathcal{F}^{Trans.}_t$ is employed to generate the separated sound feature by taking the latent mixed audio feature and visual features as inputs:
\begin{equation}
    \begin{split}
        &\boldsymbol{f}_t^{a,1} = \mathcal{F}^{Trans.}_t(\boldsymbol{f}_t^{lat.}, \boldsymbol{f}^{o,1}_t, \boldsymbol{f}^{m,1}_t), \\
        \textit{s.t.} \quad &\boldsymbol{f}^{o,1}_t=\boldsymbol{U}^o_t(\boldsymbol{Obj.}^1),\  \boldsymbol{f}^{m,1}_t=\boldsymbol{U}^m_t(\boldsymbol{Mo.}^1),
    \end{split}
    \label{eq:transformer}
\end{equation}
where $\boldsymbol{f}_t^{a,1}$ denotes the separated sound feature of video $\boldsymbol{V}_1$; $\boldsymbol{Obj.}^1$ and $\boldsymbol{Mo.}^1$ denote the object and motion features extracted by the frozen pre-trained image and video encoders respectively from the visual signal $\boldsymbol{v}_1$ of video $\boldsymbol{V}_1$, $\boldsymbol{U}^o_t(\cdot)$ and $\boldsymbol{U}^m_t(\cdot)$ are learnable projection layers to map the object and motion features into the same dimension. Similarly, we can also obtain the separated sound feature of $\boldsymbol{V}_2$ guided by the associated visual features. 

The extracted separated sound feature and the latent mixed audio feature are combined to generate a mask. This mask is subsequently applied to the mixed audio, leading to the reconstruction of the separated sound spectrogram.
\begin{equation}
    \begin{split}
        \boldsymbol{\hat{M}}^1_t=\mathcal{F}^{AD}_t(\boldsymbol{f}_t^{lat.})\odot MLP_t(\boldsymbol{f}_t^{a,1}), \\
        \boldsymbol{\hat{M}}^2_t=\mathcal{F}^{AD}_t(\boldsymbol{f}_t^{lat.})\odot MLP_t(\boldsymbol{f}_t^{a,2}),
    \end{split}
    \label{eq:gen_mask}
\end{equation}
where $\boldsymbol{\hat{M}}^1_t$ and $\boldsymbol{\hat{M}}^2_t$ denote the predicted masks for audio signals of video $\boldsymbol{V}_1$ and $\boldsymbol{V}_2$, respectively; $\mathcal{F}^{AD}_t$ is the U-Net decoder at incremental step $t$; $MLP_t(\cdot)$ denotes a MLP module; and $\odot$ denotes channel-wise multiplication. 
The sound $\boldsymbol{s}_1$ at this incremental step can be reconstructed by applying $\boldsymbol{S}\odot\boldsymbol{\hat{M}}^1_t$
and then performing an inverse STFT to obtain the audio waveform.

\subsection{Cross-modal Similarity Distillation Constraint}
\label{subsec:constraint}

Recent studies~\cite{pian2023audio,mo2023audio} have highlighted the importance of cross-modal semantic correlation in audio-visual modeling. However, this correlation tends to diminish during subsequent incremental phases, which leads to catastrophic forgetting in our continual audio-visual sound separation task. To address this challenge, we propose a novel Cross-modal Similarity Distillation Constraint (CrossSDC) that serves two crucial purposes (1) maintaining cross-modal semantic similarity through incremental tasks, and (2) preserving previous learned semantic similarity knowledge from old tasks.

CrossSDC preserves cross-modal semantic similarity from two perspectives: instance-aware semantic similarity and class-aware semantic similarity. Both similarities are enforced by integrating contrastive loss and knowledge distillation. Instead of exclusively focusing on the similarities within current and memory data generated by the current training model, CrossSDC incorporates the cross-modal similarity knowledge acquired from previous tasks into the contrastive loss. This integration not only facilitates the learning of cross-modal semantic similarities in new tasks but also ensures the preservation of previously acquired knowledge. In the incremental step $t$ ($t>1$), the instance-aware part of our CrossSDC can be formulated as: 
\begin{equation}
    \begin{split}
        &\mathcal{L}_{inst.}=-\mathbb{E}_{\boldsymbol{V}^i\sim\mathcal{D}'_t}\left[\frac{1}{\sum_j\mathbbm{1}[i=j]}\sum_j\mathbbm{1}[i=j]\log\frac{\exp(\text{sim}(\boldsymbol{f}^{mod_1}_{\tau_1,i},\boldsymbol{f}^{mod_2}_{\tau_2,j}))}{\sum_k\exp(\text{sim}(\boldsymbol{f}^{mod_1}_{\tau_1,i},\boldsymbol{f}^{mod_2}_{\tau_2,k}))}
        \right], \\
    \end{split}
    \label{eq:loss_crosssdc_ins}
\end{equation}
where $\mathbbm{1}[i=j]$ is an indicator that equals 1 when $i=j$, denoting that video samples $\boldsymbol{V}^i$ and $\boldsymbol{V}^j$ are the same video; The $\text{sim}$ function represents the cosine similarity function with temperature scaling; The modalities $mod_1$ and $mod_2$, where $(mod_1, mod_2)\in \{(a,o),(a,m),(m,o)\}$, denote different pairs of features to be compared: separated sound and object features, sound and motion features, and motion and object features. Here, $\tau$ denotes the incremental step, for which we have:
\begin{equation}
    \tau_1,\tau_2\in\textbf{T},\ where \ \textbf{T}=\left\{
    \begin{array}{ll}
        \{t,t-1\}, & \text{if } \boldsymbol{V} \in \mathcal{M}_t, \\
        \{t\}, & \text{if } \boldsymbol{V} \in \mathcal{D}_t, \\
    \end{array} \right.
\end{equation}
which means that, for current task's data $\mathcal{D}_t$, we calculate the contrastive loss using features from the current model ($\tau_1=\tau_2=t$), while for memory set data $\mathcal{M}_t$, we use features from \textit{both the old and current models} (\emph{e.g.,} $\tau_1 = t$ and $\tau_2=t-1$). In this way, knowledge distillation would be integrated into the cross-modal semantic similarity constraint for the current task, which ensures better preservation of learned cross-modal semantic similarity from previous tasks.

While the instance-aware similarity provides valuable semantic correlation modeling, it does not account for the class-level semantic correlations, which is also crucial for audio-visual similarity modeling. To capture and preserve the semantic similarity within each class across incremental tasks, we also incorporate a class-aware component specifically designed for inter-class cross-modal semantic similarity, 
which can be formulated as:
\begin{equation}
    \begin{split}
        &\mathcal{L}_{cls.}=-\mathbb{E}_{(\boldsymbol{V}^i, y^i)\sim\mathcal{D}'_t}\left[\frac{1}{\sum_j\mathbbm{1}[y^i=y^j]}\sum_j\mathbbm{1}[y^i=y^j]\log\frac{\exp(\text{sim}(\boldsymbol{f}^{mod_1}_{\tau_1,i},\boldsymbol{f}^{mod_2}_{\tau_2,j}))}{\sum_k\exp(\text{sim}(\boldsymbol{f}^{mod_1}_{\tau_1,i},\boldsymbol{f}^{mod_2}_{\tau_2,k}))}
        \right].
    \end{split}
    \label{eq:loss_crosssdc_class}
\end{equation}
In this context, visual and audio features from two videos are encouraged to be close when they belong to the same class.
The overall formulation of our CrossSDC is as follows:
\begin{equation}
    \mathcal{L}_{CrossSDC} = \lambda_{ins}\mathcal{L}_{ins}+\lambda_{cls}\mathcal{L}_{cls},
    \label{eq:loss_crosssdc}
\end{equation}
where $\lambda_{ins}$ and $\lambda_{cls}$ are two scalars that balance the two loss terms. In this way, the model captures and preserves semantic correlations not just between instances but also within the same classes.

\subsection{Overall Loss Function}
\label{subsec:loss}
In the previous subsection, we introduced our proposed CrossSDC constraint. To effectively combine CrossSDC with the overall objective, we incorporate it alongside output distillation and the main separation loss function.

Output distillation is a widely used technique in continual learning~\cite{li2017learning,ahn2021ssil,pian2023audio} to preserve the knowledge gained from previous tasks while learning new ones. In our approach, we utilize the output of the old model as the distillation target to preserve this knowledge. Note that we only distill knowledge for data from the memory set, as represented by:
\begin{equation}
    \begin{split}
        \mathcal{L}_{dist.} = \mathbb{E}_{(\boldsymbol{V}_1^i,\boldsymbol{V}_2^i)\sim\mathcal{M}_t}\Big[||\boldsymbol{\hat{M}}_{t}^1-\boldsymbol{\hat{M}}_{t-1}^1||_1 + ||\boldsymbol{\hat{M}}_{t}^2-\boldsymbol{\hat{M}}_{t-1}^2||_1\Big],
        \label{eq:loss_out_dist}
    \end{split}
\end{equation}
where $\boldsymbol{\hat{M}}_{t-1}^1$ and $\boldsymbol{\hat{M}}_{t-1}^2$ are predicted masks generated by the old model that is trained at incremental step $t-1$. For the loss function here, we follow~\cite{zhao2018sound,chen2023iquery} and adopt the per-pixel $L_1$ loss~\cite{zhao2018sound}. For the main separation loss function, we also apply the per-pixel $L_1$ loss:
\begin{equation}
    \begin{split}
        \mathcal{L}_{main} = \mathbb{E}_{(\boldsymbol{V}_1^i,\boldsymbol{V}_2^i)\sim\mathcal{M}_t}\Big[||\boldsymbol{\hat{M}}_{t}^1-\boldsymbol{M}^1||_1 + ||\boldsymbol{\hat{M}}_{t}^2-\boldsymbol{M}^2||_1\Big],
        \label{eq:loss_main}
    \end{split}
\end{equation}
Finally, our overall loss function is denoted as:
\begin{equation}
    \mathcal{L}_{ContAV-Sep} = \mathcal{L}_{main} + \lambda_{dist.}\mathcal{L}_{dist.} + \mathcal{L}_{CrossSDC}.
    \label{eq:loss_final}
\end{equation}

\subsection{Management of Memory Set}
\label{subsec:memory}

In alignment with the work of~\cite{wang2022learning}, our framework maintains a compact memory set throughout incremental updates. Each old class is limited to a maximum number of exemplars. After completing training for each task, we adopt the exemplar selection strategies in \cite{ahn2021ssil,pian2023audio} by randomly selecting exemplars for each current class and combining these new exemplars with the existing memory set.

\section{Experiments}

In this section, we first introduce the setup of our experiments, \textit{i.e.}, dataset, baselines, evaluation metrics, and the implementation details. After that, we present the experimental results of our ContAV-Sep compared to the baselines, as well as ablation studies. We also conduct experiments on the AVE~\cite{tian2018audio} and the VGGSound~\cite{chen2020vggsound} datasets, which contain sound categories beyond the music domain. 
We put the experimental results on the AVE and the VGGSound datasets, the comparison to the uni-modal semantic similarity preservation method, the performance evaluation on old classes in incremental tasks, and the visualization of separating results in the Appendix.


\subsection{Experimental Setup}
\label{sec:exp_setup}

\textbf{Dataset.}
Following common practice~\cite{zhao2019sound,zhu2022visually,chen2023iquery}, we conducted experiments on \textit{MUSIC-21}~\cite{zhao2019sound}, which contains solo videos of 21 instruments categories: accordion, acoustic guitar, cello, clarinet, erhu, flute, saxophone, trumpet, tuba, violin, xylophone, bagpipe, banjo, bassoon, congas, drum, electric, bass, guzheng, piano, pipa, and ukulele. In our experiments, we randomly selected 20 of them to construct the continual learning setting. Specifically, we split the selected 20 classes into 4 incremental tasks, each of which involves 5 classes. The total number of available videos is 1040, and we randomly split them into training, validation, and testing sets with 840, 100, and 100 videos, respectively. To further validate the efficacy of our method across a broader sound domain, we conduct experiments using the AVE~\cite{tian2018audio} and the VGGSound~\cite{chen2020vggsound} datasets in the appendix.

\textbf{Baselines.}
We compare our proposed approach with vanilla Fine-tuning strategy, and continual learning methods EWC~\cite{kirkpatrick2017overcoming} and LwF~\cite{li2017learning}. As we mentioned before, typical continual learning methods, \textit{e.g.}, class-incremental learning methods, which yield progressively increasing logits (or probability distribution) across all observed classes at each incremental step and design specific technique in the classifier, we consider that these methods are not an optimal choice for our proposed continual audio-visual sound separation problem. Thus, considering that continual semantic segmentation problem has a more similar form compared to conventional class-incremental learning, we also select two state-of-the-art continual semantic segmentation methods PLOP~\cite{douillard2021plop} and EWF~\cite{xiao2023endpoints} as our baselines. 
Moreover, we compare our method to the recently proposed audio-visual continual learning method AV-CIL~\cite{pian2023audio}, in which we adapt the original class-incremental version to the form of continual audio-visual sound separation by replacing their task-wise logits distillation with the output mask distillation.
Further, we also present the experimental results of the Oracle/Upper Bound, which means that using the training data from all seen classes to train the model. \textbf{For fair comparison, all compared continual learning methods and our ContAV-Sep use the same state-of-the-art separator, \textit{i.e.} iQuery~\cite{chen2023iquery}, as the base separation model}. Further, we also incorporate our proposed and baseline methods into another representative audio-visual sound separation model Co-Separation~\cite{gao2019co}. Notably, the Co-Separation model does not utilize the motion modality. Therefore, when CrossSDC is applied to Co-Separation, the $(mod_1, mod_2)$ in Eq.~\ref{eq:loss_crosssdc_ins} and~\ref{eq:loss_crosssdc_class} is constrained to $(mod_1, mod_2)=(a,o)$.
For baselines that involve memory sets, we ensure that each of them is allocated the same number of memory as our proposed method for fair comparison. 

\noindent
\textbf{Implementation Details.}
Following~\cite{chen2023iquery}, we use a 7-layers U-Net~\cite{ronneberger2015u} as the audio net, and sub-sample the audio at 11kHz, each of which is approximately 6 seconds. We apply the STFT with the Hann window size of 1022 and the hop length of 256, to obtain the $512\times 256$ Time-Frequency representation of each audio signal, followed by a re-sampling on the log-frequency scale to generate the magnitude spectrogram with $T,F=256$. We set the video frame rate (FPS) to 1, and detect the object using the pre-trained universal detector Detic~\cite{zhou2022detecting} to detect the sound source object on each frame, and then, each detected object is resized and randomly cropped to the size of $224\times 224$. For the image encoder and the video encoder, we apply the self-supervised pre-trained CLIP~\cite{radford2021learning} and VideoMAE~\cite{tong2022videomae} to yield the object feature and motion feature, respectively. For the audio-visual Transformer module, we follow the design in~\cite{chen2023iquery}. For all the baseline methods, we apply the same model architecture and modules with ours for them, including the mentioned Detic, CLIP, VideoMAE, audio-visual Transformer, etc. 
Please note that, during our training process, the pre-trained Detic, CLIP, and VideoMAE are frozen. In our proposed Cross-modal Similarity Distillation Constraint (CrossSDC), the balance weights $\lambda_{ins}$ and $\lambda_{cls}$ are set to 0.1 and 0.3, respectively. And the balance weight $\lambda_{dist.}$ for the output distillation loss is set to 0.3 in our experiments. For the memory set, we set the number of samples in each old class to 1, so as other baselines that involve the memory set. All the experiments in this paper are implemented by Pytorch~\cite{paszke2019pytorch}. We train our proposed method and all baselines on a NVIDIA RTX A5000 GPU.
We follow previous works~\cite{tian2021cyclic,chen2023iquery} in sound separation, and evaluate the performance of all the methods using three common metrics in sound separation tasks: Signal to Distortion Ratio (SDR), Signal to Interference Ratio (SIR), and Signal to Artifact Ratio (SAR). The SDR measures the interference and artifacts, while SIR and SAR measure the interference and artifacts, respectively. In our experiments, we report the SDR, SIR, and SAR of all the methods after training at last incremental steps, \textit{i.e.}, testing results on all classes. For all these three metrics, higher values denote better results.

\begin{table}[t]
  \centering

  \caption{Main results of different methods on MUSIC-21 dataset under the setting of Continual Audio-Visual Sound Separation with base separation models of iQuery~\cite{chen2023iquery} and Co-Separation~\cite{gao2019co}, respectively. The bold part denotes the best results. Our proposed ContAV-Sep achieves the best performance among all baselines.}
  \resizebox{\textwidth}{!}{
    \begin{tabular}{lccc|lccc}
    \toprule
    Method & SDR$\uparrow$   & SIR$\uparrow$   & SAR$\uparrow$   & Method & SDR$\uparrow$   & SIR$\uparrow$   & SAR$\uparrow$ \\
    \midrule
    \multicolumn{4}{l|}{\textit{w/o memory}} & \multicolumn{4}{l}{\textit{w/o memory}} \\
    iQuery~\cite{chen2023iquery} + Fine-tuning & 3.46  & 9.30  & 10.57 & Co-Sep.~\cite{gao2019co} + Fine-tuning & 1.93  & 8.75  & 9.75 \\
    iQuery~\cite{chen2023iquery} + LwF~\cite{li2017learning} & 3.45  & 8.78  & 10.66 & Co-Sep.~\cite{gao2019co} + LwF~\cite{li2017learning} & 2.32  & 7.84  & 10.28 \\
    iQuery~\cite{chen2023iquery} + EWC~\cite{lee2019overcoming} & 3.67  & 9.58  & 10.30 & Co-Sep.~\cite{gao2019co} + EWC~\cite{lee2019overcoming} & 2.01  & 8.36  & 9.61 \\
    iQuery~\cite{chen2023iquery} + PLOP~\cite{douillard2021plop} & 3.82  & 10.06 & 10.22 & Co-Sep.~\cite{gao2019co} + PLOP~\cite{douillard2021plop} & 3.24  & 9.17  & 9.59 \\
    iQuery~\cite{chen2023iquery} + EWF~\cite{xiao2023endpoints} & 3.98  & 9.68  & 11.52 & Co-Sep.~\cite{gao2019co} + EWF~\cite{xiao2023endpoints} & 2.61  & 7.77  & 10.85 \\
    \midrule
    \textit{w/ memory} &       &       &       & \textit{w/ memory} &       &       &  \\
    iQuery~\cite{chen2023iquery} + LwF~\cite{li2017learning} & 6.76  & 12.77 & 12.60 & Co-Sep.~\cite{gao2019co} + LwF~\cite{li2017learning} & 3.85  & 9.62  & 10.74 \\
    iQuery~\cite{chen2023iquery} + EWC~\cite{lee2019overcoming} & 6.65  & 13.01 & 11.73 & Co-Sep.~\cite{gao2019co} + EWC~\cite{lee2019overcoming} & 3.31  & 9.55  & 9.80 \\
    iQuery~\cite{chen2023iquery} + PLOP~\cite{douillard2021plop} & 7.03  & 13.30 & 11.90 & Co-Sep.~\cite{gao2019co} + PLOP~\cite{douillard2021plop} & 3.88  & 9.92  & 9.99 \\
    iQuery~\cite{chen2023iquery} + EWF~\cite{xiao2023endpoints} & 5.35  & 11.35 & 11.81 & Co-Sep.~\cite{gao2019co} + EWF~\cite{xiao2023endpoints} & 3.63  & 9.07  & 10.58 \\
    iQuery~\cite{chen2023iquery} + AV-CIL~\cite{pian2023audio} & 6.86 & 13.13 & 12.31 & Co-Sep.~\cite{gao2019co} + AV-CIL~\cite{pian2023audio} & 3.61 & 9.76 & 9.68 \\
    \textbf{ContAV-Sep (with iQuery~\cite{chen2023iquery})} & \textbf{7.33} & \textbf{13.55} & \textbf{13.01} & \textbf{ContAV-Sep (with Co-Sep.~\cite{gao2019co})} & \textbf{4.06} & \textbf{10.06} & \textbf{11.07} \\
    \midrule
    Upper Bound (with iQuery) & 10.36 & 16.64 & 14.68 & Upper Bound (with Co-Sep.) & 7.30  & 14.34 & 11.90 \\
    \bottomrule
    \end{tabular}%
    }
  \label{tab:main_res}%
  \vspace{-5mm}
\end{table}%

\subsection{Experimental Comparison}
The main experimental comparisons are shown in Tab.~\ref{tab:main_res}. Our proposed method, ContAV-Sep, outperforms the state-of-the-art baselines by a substantial margin. Notably, compared to baselines using state-of-the-art audio-visual sound separator iQuery~\cite{chen2023iquery} as the separation base model, ContAV-Sep achieves a {0.3} improvement in SDR over the compared best-performing method. Additionally, our method surpasses the top baseline by {0.25} in SIR and {0.41} in SAR. Furthermore, compared to continual learning baselines with Co-Separation~\cite{gao2019co}, our ContAV-Sep still outperforms other approaches. This consistent superior performance across different model architectures highlights not only the effectiveness but also the broad applicability and generalizability of our proposed CrossSDC.


Our observations further demonstrate that retaining a small memory set significantly enhances the performance of each baseline method. For instance, for the iQuery-based continual learning methods, equipping LwF~\cite{li2017learning} with a small memory set results in improvements of 3.31, 3.99, and 1.94 on SDR, SIR, and SAR, respectively. Similarly, the addition of a small memory set to EWC~\cite{kirkpatrick2017overcoming} leads to enhancements of 2.98, 3.43, and 1.43 in the respective metrics. The memory-augmented version of PLOP~\cite{douillard2021plop} exhibits superior performance with margins of 3.21, 3.24, and 1.68 for SDR, SIR, and SAR, respectively. Finally, incorporating memory into EWF~\cite{xiao2023endpoints} results in improvements of 1.37, 1.67, and 0.29 for the three metrics.
This phenomenon can be attributed to the inherent nature of the sound separation training process. In training, the audio signal from each sample mixes with others, giving a composite audio signal. This mixed audio signal, coupled with the corresponding visual data pair for each separated audio, constitutes the actual training sample for the separation task. As a result, even a single memory sample can be associated with multiple samples from the current training set, generating a diverse array of effective training pairs.

\begin{figure*}[t]
    \centering
    \subfloat[]{
    \centering
    \includegraphics[width=0.32\textwidth,height=0.25\textwidth]{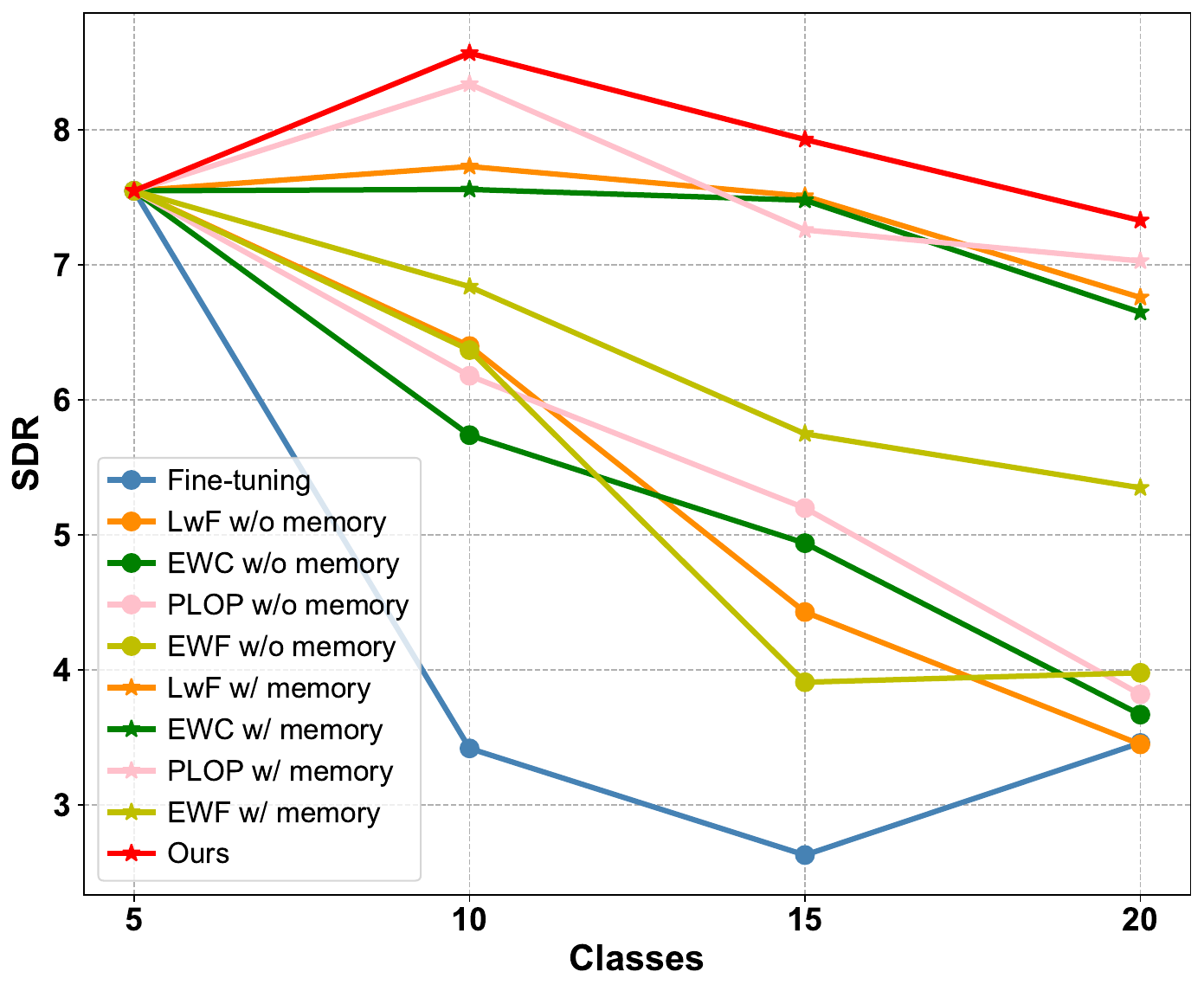}
    \label{fig:step_sdr}
    }
    \subfloat[]{
    \centering
    \includegraphics[width=0.32\textwidth,height=0.25\textwidth]{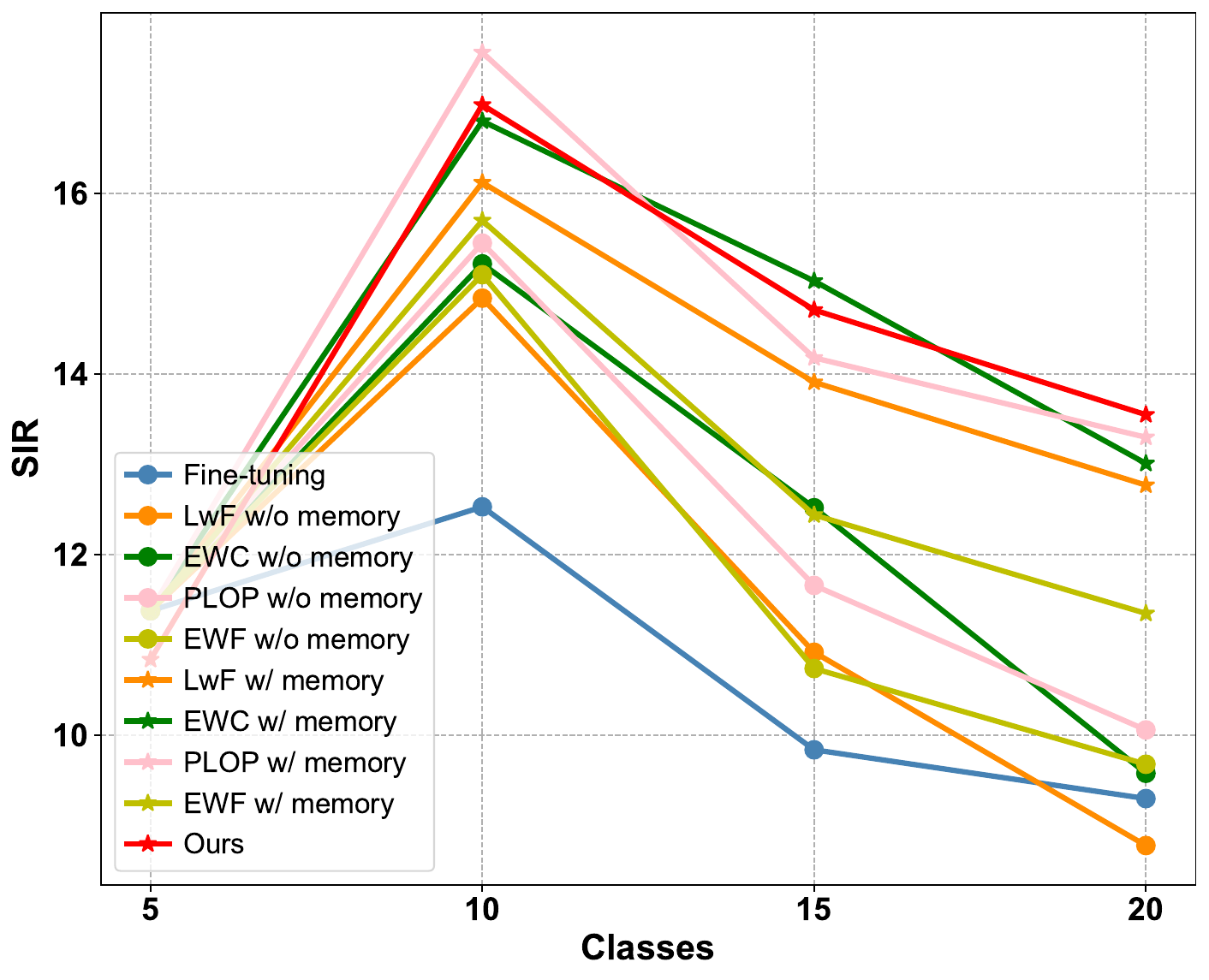}
    \label{fig:step_sir}
    }
    \subfloat[]{
    \centering
    \includegraphics[width=0.32\textwidth,height=0.25\textwidth]{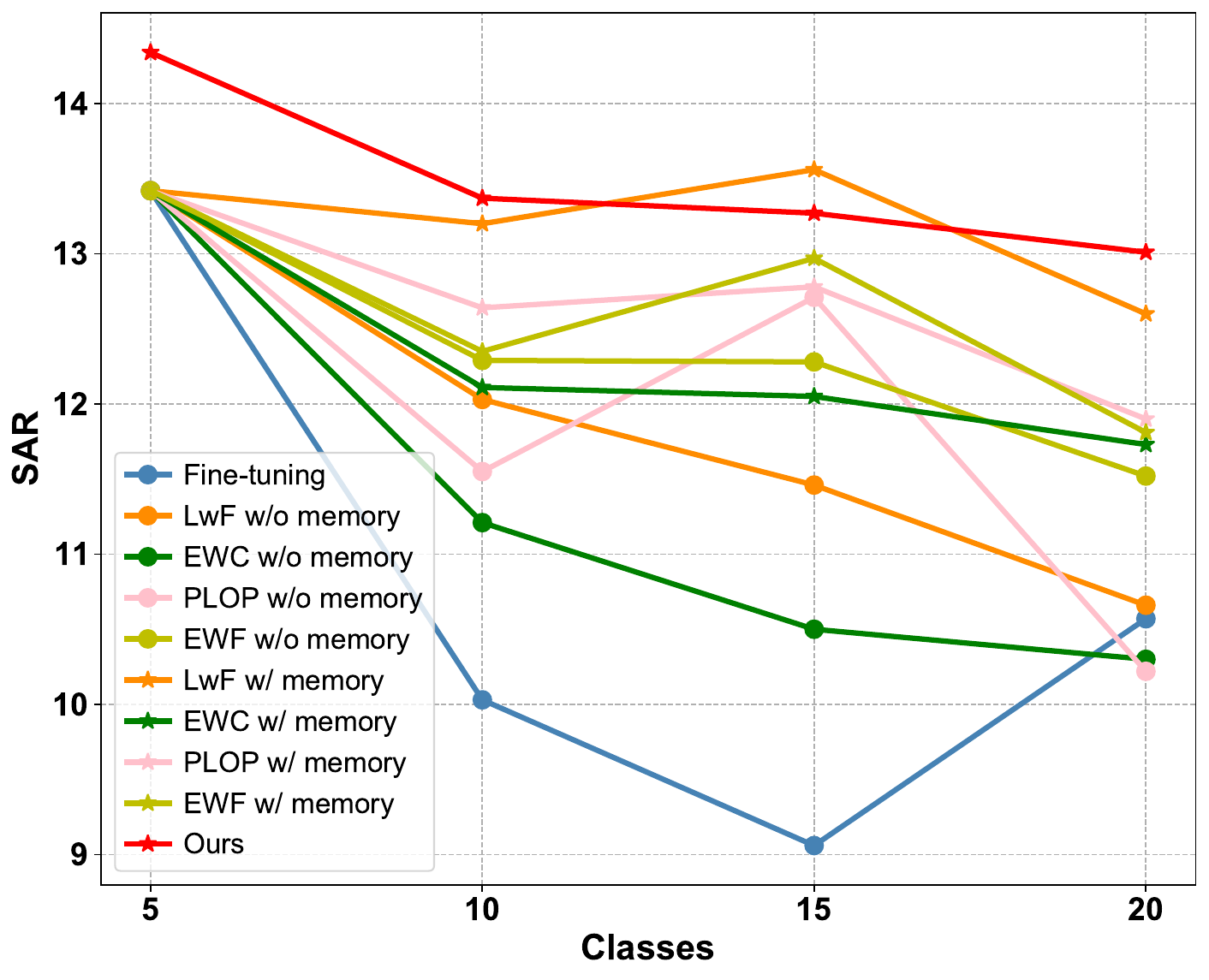}
    \label{fig:step_sar}
    }
    \vspace{-1mm}
    \caption{Testing results of different continual learning methods with iQuery~\cite{chen2023iquery} on the metrics of (a) SDR, (b) SIR, and (c) SAR at each incremental step.}
\vspace{-4mm}
\end{figure*}


We also present the testing results of SDR, SIR, and SAR at each incremental step in Figures~\ref{fig:step_sdr},~\ref{fig:step_sir}, and~\ref{fig:step_sar}, respectively. 
Our method is consistently observed to outperform others in terms of SDR at all incremental steps.
While our approach may not always produce the best SIR and SAR results at the intermediate steps (specifically, steps 2 and 3 for SIR, and step 3 for SAR), it ultimately achieves the highest performance at the final step. This demonstrates the robustness of our method, indicating minimal forgetting throughout the incremental learning process.



\subsection{Ablation Study on CrossSDC and Memory Size}
In this subsection, we conduct an ablation study to investigate the effectiveness of our proposed CrossSDC. By removing single or multiple components of the CrossSDC, we evaluate the impact of each on the final results. The results of the ablation study are presented in Tab.~\ref{tab:ablation}. From the table, we can see that our full model achieves the best performance compared to the variants, which further demonstrates the effectiveness of our proposed CrossSDC.

\begin{figure*}[t]
    \centering
    \subfloat[]{
    \centering
    \includegraphics[width=0.32\textwidth,height=0.25\textwidth]{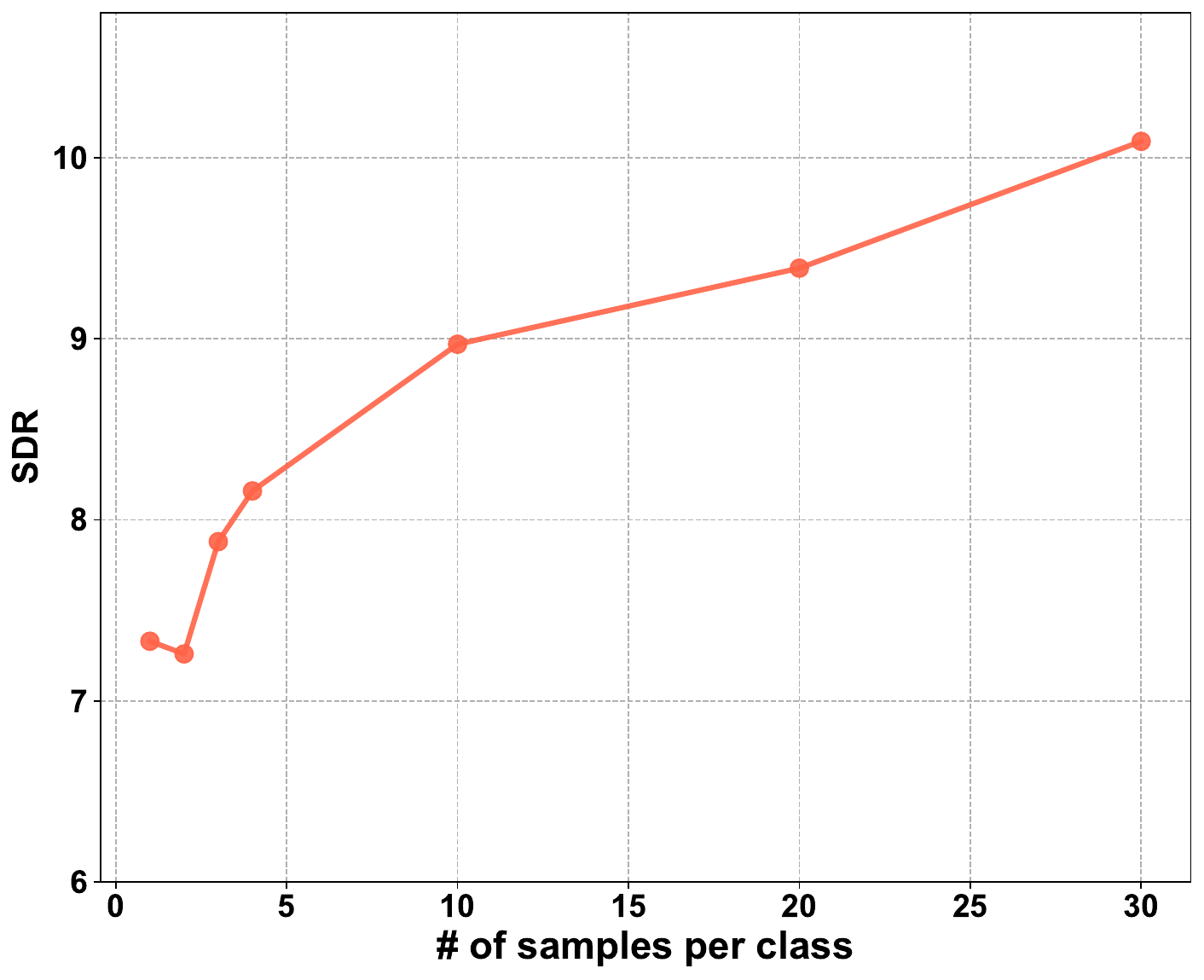}
    \label{fig:memory_sdr}
    }
    \subfloat[]{
    \centering
    \includegraphics[width=0.32\textwidth,height=0.25\textwidth]{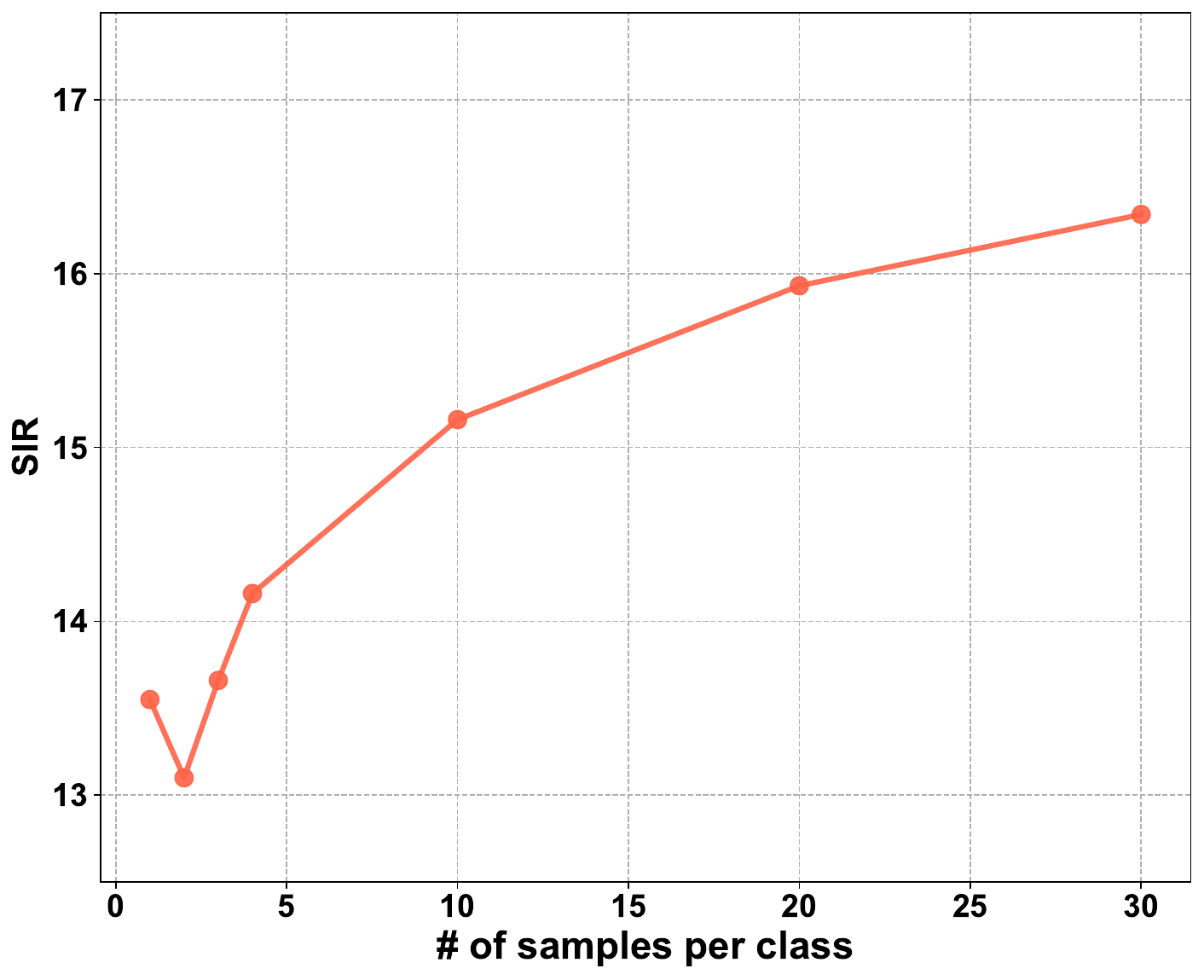}
    \label{fig:memory_sir}
    }
    \subfloat[]{
    \centering
    \includegraphics[width=0.32\textwidth,height=0.25\textwidth]{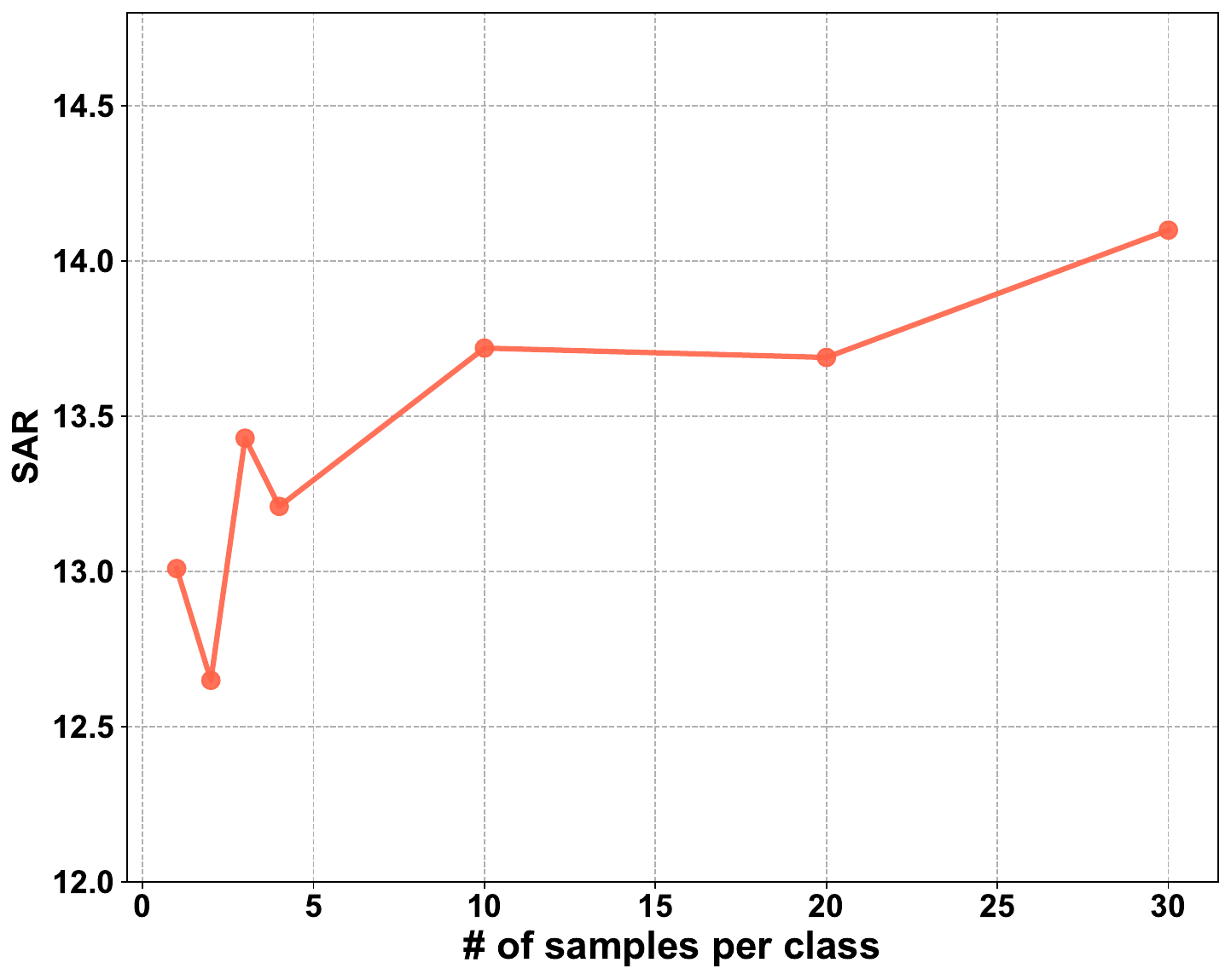}
    \label{fig:memory_sar}
    }
    
    \vspace{-1mm}
    \caption{Testing results with different memory size (number of samples per class in the memory) on the metrics of (a) SDR, (b) SIR, and (c) SAR at each incremental step.}
    \label{fig:memory_size}
\vspace{-4mm}
\end{figure*}

Moreover, we also discuss the effect of memory size on our proposed ContAV-Sep. In our main experiments, the default setting of the memory size is 1 sample per old class. In this subsection, we conduct experiments by increasing the memory size from 1 sample per old class to 30 samples per old class. The experimental results are shown in Tab.~\ref{tab:memory_size} and Figure~\ref{fig:memory_size}. Observations from the table indicate a positive correlation between the size of the memory and the overall performance metrics. As the memory size increases, there is a discernible trend of improvement in the results.

\begin{table}[htbp]
\vspace{-3mm}
  \centering
  \caption{Ablation study on our proposed ContAV-Sep. Our full approach achieves best results compared to the variants.}
    \begin{tabular}{lcccccc}
    \toprule
    \multirow{5}[4]{*}{ContAV-Sep} & $\mathcal{L}_{dist.}$ & $\mathcal{L}_{inst.}$ & $\mathcal{L}_{cls.}$ & SDR$\uparrow$   & SIR$\uparrow$   & SAR$\uparrow$ \\
\cmidrule{2-7}          & \CheckmarkBold     & \XSolidBrush     & \XSolidBrush     &   6.32    &   12.99    & 11.82 \\
          & \CheckmarkBold     & \CheckmarkBold     & \XSolidBrush     &   6.01    &   11.92  &   11.74   \\
          & \CheckmarkBold     & \XSolidBrush     & \CheckmarkBold     &   6.86   &  13.12   &  12.25 \\
          & \CheckmarkBold     & \CheckmarkBold     & \CheckmarkBold     &   \textbf{7.33}    &   \textbf{13.55}  &    \textbf{13.01}   \\
    \bottomrule
    \end{tabular}%
  \label{tab:ablation}%
\vspace{-3mm}
\end{table}%

\begin{table}[htbp]
  \centering
  \caption{Experimental results of our proposed ContAV-Sep with different memory size from 1 to 30 samples per memory class.}
    \begin{tabular}{lcccc}
    \toprule
    \multirow{8}[4]{*}{ContAV-Sep} & \# of samples per class & SDR$\uparrow$   & SIR$\uparrow$   & SAR$\uparrow$ \\
\cmidrule{2-5}          & 1     & 7.33  & 13.55 & 13.01 \\
          & 2     & 7.26  & 13.10 & 12.65 \\
          & 3     & 7.88  & 13.66 & 13.43 \\
          & 4     & 8.16  & 14.16 & 13.21 \\
          & 10    & 8.97  & 15.16 & 13.72 \\
          & 20    & 9.39  & 15.93 & 13.69 \\
          & 30    & \textbf{10.09} & \textbf{16.34} & \textbf{14.10} \\
    \bottomrule
    \end{tabular}%
  \label{tab:memory_size}%
\end{table}%

\subsection{Limitation and Discussion}
\label{sec:limitation}

Our experimental findings reveal that the utilization of a small memory set, even a single sample per old class, markedly improves the performance of each baseline method. This improvement is attributed to the ability of a single memory sample to pair with diverse samples from the current training set, thereby generating numerous effective training pairs. Consequently, this process enables the model to acquire new knowledge for old classes in subsequent tasks, as the memory data can be paired with data from previously unseen new classes --- this is different from conventional continual learning tasks, where old classes do \textit{not} acquire new knowledge in new tasks. This could be a potential reason why the baseline continual learning methods do not perform well in our continual audio-visual sound separation problem. In this work, our method also mainly focuses on preserving old knowledge of old tasks, which may prevent the model from acquiring new knowledge of old classes when training in new tasks. Recognizing this, we identify the exploration of this problem as a key avenue for future research in this field.

Additionally, the base model architectures used in our approach and baselines require object detectors to identify sounding objects.
Although iQuery~\cite{chen2023iquery} can supplement object features with global video representations, it may still suffer from undetected objects. It is a fundamental limitation of the \textbf{object-based audio-visual sound separators}~\cite{gao2019co,chen2023iquery}. 
While our work, unlike previous efforts, does not compete on designing a stronger audio-visual separation base model, enhancing the robustness of sounding object detection presents a promising direction for future research.

\section{Conclusion}
\label{sec:conclusion}

In this paper, we explore training audio-visual sound separation models under a more practical continual learning scenario, and introduce the task of continual audio-visual sound separation. To address this novel problem, we propose ContAV-Sep, which incorporates a Cross-modal Similarity Distillation Constraint to maintain cross-modal semantic similarity across incremental tasks while preserving previously learned semantic similarity knowledge. Experiments on the MUSIC-21 dataset demonstrate the effectiveness of our method in this new continual separation task. This paper opens a new direction for real-world audio-visual sound separation research.

\textbf{Broader Impact.} Our proposed continual audio-visual sound separation allows the model to adapt to new environments and sounds without full retraining, which could enhance efficiency and privacy by reducing the need to transmit and store sensitive audio data.

\textbf{Acknowledgments.} We thank the anonymous reviewers and area chair for their valuable suggestions and comments. This work was supported in part by
a Cisco Faculty Research Award, an Amazon Research
Award, and a research gift from Adobe. The article solely
reflects the opinions and conclusions of its authors but not
the funding agents.

\bibliographystyle{plain}
\bibliography{reference}

\begin{thebibliography}{10}

\bibitem{afouras2018conversation}
Triantafyllos Afouras, Joon~Son Chung, and Andrew Zisserman.
\newblock The conversation: Deep audio-visual speech enhancement.
\newblock {\em arXiv preprint arXiv:1804.04121}, 2018.

\bibitem{ahn2021ssil}
Hongjoon Ahn, Jihwan Kwak, Subin Lim, Hyeonsu Bang, Hyojun Kim, and Taesup Moon.
\newblock {SS-IL:} separated softmax for incremental learning.
\newblock In {\em Proceedings of the IEEE/CVF International Conference on Computer Vision}, pages 824--833, 2021.

\bibitem{aljundi2018memory}
Rahaf Aljundi, Francesca Babiloni, Mohamed Elhoseiny, Marcus Rohrbach, and Tinne Tuytelaars.
\newblock Memory aware synapses: Learning what (not) to forget.
\newblock In {\em Proceedings of the European conference on computer vision (ECCV)}, pages 139--154, 2018.

\bibitem{arani2022learning}
Elahe Arani, Fahad Sarfraz, and Bahram Zonooz.
\newblock Learning fast, learning slow: A general continual learning method based on complementary learning system.
\newblock {\em arXiv preprint arXiv:2201.12604}, 2022.

\bibitem{belouadah2019il2m}
Eden Belouadah and Adrian Popescu.
\newblock Il2m: Class incremental learning with dual memory.
\newblock In {\em Proceedings of the IEEE/CVF international conference on computer vision}, pages 583--592, 2019.

\bibitem{bhatt2024characterizing}
Ruchi Bhatt, Pratibha Kumari, Dwarikanath Mahapatra, Abdulmotaleb~El Saddik, and Mukesh Saini.
\newblock Characterizing continual learning scenarios and strategies for audio analysis.
\newblock {\em arXiv preprint arXiv:2407.00465}, 2024.

\bibitem{bregnian1993auditory}
Albert~S Bregnian.
\newblock Auditory scene analysis: Hearing in complex environments.
\newblock {\em Thinking in Sounds}, pages 10--36, 1993.

\bibitem{buzzega2020dark}
Pietro Buzzega, Matteo Boschini, Angelo Porrello, Davide Abati, and Simone Calderara.
\newblock Dark experience for general continual learning: a strong, simple baseline.
\newblock {\em Advances in neural information processing systems}, 33:15920--15930, 2020.

\bibitem{castro2018end}
Francisco~M Castro, Manuel~J Mar{\'\i}n-Jim{\'e}nez, Nicol{\'a}s Guil, Cordelia Schmid, and Karteek Alahari.
\newblock End-to-end incremental learning.
\newblock In {\em Proceedings of the European Conference on Computer Vision (ECCV)}, pages 233--248, 2018.

\bibitem{cha2023rebalancing}
Sungmin Cha, Sungjun Cho, Dasol Hwang, Sunwon Hong, Moontae Lee, and Taesup Moon.
\newblock Rebalancing batch normalization for exemplar-based class-incremental learning.
\newblock In {\em Proceedings of the IEEE/CVF Conference on Computer Vision and Pattern Recognition}, pages 20127--20136, 2023.

\bibitem{chatterjee2022learning}
Moitreya Chatterjee, Narendra Ahuja, and Anoop Cherian.
\newblock Learning audio-visual dynamics using scene graphs for audio source separation.
\newblock {\em Advances in Neural Information Processing Systems}, 35:16975--16988, 2022.

\bibitem{chaudhry2019tiny}
Arslan Chaudhry, Marcus Rohrbach, Mohamed Elhoseiny, Thalaiyasingam Ajanthan, Puneet~K Dokania, Philip~HS Torr, and Marc'Aurelio Ranzato.
\newblock On tiny episodic memories in continual learning.
\newblock {\em arXiv preprint arXiv:1902.10486}, 2019.

\bibitem{chen2020vggsound}
Honglie Chen, Weidi Xie, Andrea Vedaldi, and Andrew Zisserman.
\newblock Vggsound: A large-scale audio-visual dataset.
\newblock In {\em ICASSP 2020-2020 IEEE International Conference on Acoustics, Speech and Signal Processing (ICASSP)}, pages 721--725. IEEE, 2020.

\bibitem{chen2023iquery}
Jiaben Chen, Renrui Zhang, Dongze Lian, Jiaqi Yang, Ziyao Zeng, and Jianbo Shi.
\newblock iquery: Instruments as queries for audio-visual sound separation.
\newblock In {\em Proceedings of the IEEE/CVF Conference on Computer Vision and Pattern Recognition}, pages 14675--14686, 2023.

\bibitem{chung2020facefilter}
Soo-Whan Chung, Soyeon Choe, Joon~Son Chung, and Hong-Goo Kang.
\newblock Facefilter: Audio-visual speech separation using still images.
\newblock {\em arXiv preprint arXiv:2005.07074}, 2020.

\bibitem{douillard2021plop}
Arthur Douillard, Yifu Chen, Arnaud Dapogny, and Matthieu Cord.
\newblock Plop: Learning without forgetting for continual semantic segmentation.
\newblock In {\em Proceedings of the IEEE/CVF conference on computer vision and pattern recognition}, pages 4040--4050, 2021.

\bibitem{douillard2020podnet}
Arthur Douillard, Matthieu Cord, Charles Ollion, Thomas Robert, and Eduardo Valle.
\newblock Podnet: Pooled outputs distillation for small-tasks incremental learning.
\newblock In {\em Computer Vision--ECCV 2020: 16th European Conference, Glasgow, UK, August 23--28, 2020, Proceedings, Part XX 16}, pages 86--102. Springer, 2020.

\bibitem{ephrat2018looking}
Ariel Ephrat, Inbar Mosseri, Oran Lang, Tali Dekel, Kevin Wilson, Avinatan Hassidim, William~T Freeman, and Michael Rubinstein.
\newblock Looking to listen at the cocktail party: A speaker-independent audio-visual model for speech separation.
\newblock {\em arXiv preprint arXiv:1804.03619}, 2018.

\bibitem{fini2022self}
Enrico Fini, Victor G~Turrisi Da~Costa, Xavier Alameda-Pineda, Elisa Ricci, Karteek Alahari, and Julien Mairal.
\newblock Self-supervised models are continual learners.
\newblock In {\em Proceedings of the IEEE/CVF Conference on Computer Vision and Pattern Recognition}, pages 9621--9630, 2022.

\bibitem{gabbay2017visual}
Aviv Gabbay, Asaph Shamir, and Shmuel Peleg.
\newblock Visual speech enhancement.
\newblock {\em arXiv preprint arXiv:1711.08789}, 2017.

\bibitem{gan2020music}
Chuang Gan, Deng Huang, Hang Zhao, Joshua~B Tenenbaum, and Antonio Torralba.
\newblock Music gesture for visual sound separation.
\newblock In {\em Proceedings of the IEEE/CVF Conference on Computer Vision and Pattern Recognition}, pages 10478--10487, 2020.

\bibitem{gao2018learning}
Ruohan Gao, Rogerio Feris, and Kristen Grauman.
\newblock Learning to separate object sounds by watching unlabeled video.
\newblock In {\em Proceedings of the European Conference on Computer Vision (ECCV)}, pages 35--53, 2018.

\bibitem{gao2019co}
Ruohan Gao and Kristen Grauman.
\newblock Co-separating sounds of visual objects.
\newblock In {\em Proceedings of the IEEE/CVF International Conference on Computer Vision}, pages 3879--3888, 2019.

\bibitem{golkar2019continual}
Siavash Golkar, Michael Kagan, and Kyunghyun Cho.
\newblock Continual learning via neural pruning.
\newblock {\em arXiv preprint arXiv:1903.04476}, 2019.

\bibitem{haykin2005cocktail}
Simon Haykin and Zhe Chen.
\newblock The cocktail party problem.
\newblock {\em Neural computation}, 17(9):1875--1902, 2005.

\bibitem{hou2019learning}
Saihui Hou, Xinyu Pan, Chen~Change Loy, Zilei Wang, and Dahua Lin.
\newblock Learning a unified classifier incrementally via rebalancing.
\newblock In {\em Proceedings of the IEEE/CVF Conference on Computer Vision and Pattern Recognition}, pages 831--839, 2019.

\bibitem{huang2023davis}
Chao Huang, Susan Liang, Yapeng Tian, Anurag Kumar, and Chenliang Xu.
\newblock Davis: High-quality audio-visual separation with generative diffusion models.
\newblock {\em arXiv preprint arXiv:2308.00122}, 2023.

\bibitem{hung2019compacting}
Ching-Yi Hung, Cheng-Hao Tu, Cheng-En Wu, Chien-Hung Chen, Yi-Ming Chan, and Chu-Song Chen.
\newblock Compacting, picking and growing for unforgetting continual learning.
\newblock {\em Advances in Neural Information Processing Systems}, 32, 2019.

\bibitem{kang2022class}
Minsoo Kang, Jaeyoo Park, and Bohyung Han.
\newblock Class-incremental learning by knowledge distillation with adaptive feature consolidation.
\newblock In {\em Proceedings of the IEEE/CVF Conference on Computer Vision and Pattern Recognition}, pages 16050--16059, 2022.

\bibitem{ke2023continual}
Zixuan Ke, Yijia Shao, Haowei Lin, Tatsuya Konishi, Gyuhak Kim, and Bing Liu.
\newblock Continual learning of language models.
\newblock In {\em International Conference on Learning Representations (ICLR)}, 2023.

\bibitem{kim2023achieving}
Sanghwan Kim, Lorenzo Noci, Antonio Orvieto, and Thomas Hofmann.
\newblock Achieving a better stability-plasticity trade-off via auxiliary networks in continual learning.
\newblock In {\em Proceedings of the IEEE/CVF Conference on Computer Vision and Pattern Recognition}, pages 11930--11939, 2023.

\bibitem{kirkpatrick2017overcoming}
James Kirkpatrick, Razvan Pascanu, Neil Rabinowitz, Joel Veness, Guillaume Desjardins, Andrei~A Rusu, Kieran Milan, John Quan, Tiago Ramalho, Agnieszka Grabska-Barwinska, et~al.
\newblock Overcoming catastrophic forgetting in neural networks.
\newblock {\em Proceedings of the National Academy of Sciences}, 114(13):3521--3526, 2017.

\bibitem{kushwaha2022analyzing}
Saksham~Singh Kushwaha.
\newblock Analyzing the effect of equal-angle spatial discretization on sound event localization and detection.
\newblock In {\em Proceedings of the 7th Detection and Classification of Acoustic Scenes and Events 2022 Workshop (DCASE2022)}, 2022.

\bibitem{kushwaha2023multimodal}
Saksham~Singh Kushwaha and Magdalena Fuentes.
\newblock A multimodal prototypical approach for unsupervised sound classification.
\newblock {\em arXiv preprint arXiv:2306.12300}, 2023.

\bibitem{lee2023lifelong}
Jaewoo Lee, Jaehong Yoon, Wonjae Kim, Yunji Kim, and Sung~Ju Hwang.
\newblock Lifelong audio-video masked autoencoder with forget-robust localized alignments.
\newblock {\em arXiv preprint arXiv:2310.08204}, 2023.

\bibitem{lee2019overcoming}
Kibok Lee, Kimin Lee, Jinwoo Shin, and Honglak Lee.
\newblock Overcoming catastrophic forgetting with unlabeled data in the wild.
\newblock In {\em Proceedings of the IEEE/CVF International Conference on Computer Vision}, pages 312--321, 2019.

\bibitem{li2019learn}
Xilai Li, Yingbo Zhou, Tianfu Wu, Richard Socher, and Caiming Xiong.
\newblock Learn to grow: A continual structure learning framework for overcoming catastrophic forgetting.
\newblock In {\em International Conference on Machine Learning}, pages 3925--3934. PMLR, 2019.

\bibitem{li2017learning}
Zhizhong Li and Derek Hoiem.
\newblock Learning without forgetting.
\newblock {\em IEEE transactions on pattern analysis and machine intelligence}, 40(12):2935--2947, 2017.

\bibitem{liang2023adaptive}
Yan-Shuo Liang and Wu-Jun Li.
\newblock Adaptive plasticity improvement for continual learning.
\newblock In {\em Proceedings of the IEEE/CVF Conference on Computer Vision and Pattern Recognition}, pages 7816--7825, 2023.

\bibitem{luo2023class}
Zilin Luo, Yaoyao Liu, Bernt Schiele, and Qianru Sun.
\newblock Class-incremental exemplar compression for class-incremental learning.
\newblock In {\em Proceedings of the IEEE/CVF Conference on Computer Vision and Pattern Recognition}, pages 11371--11380, 2023.

\bibitem{ma2021continual}
Haoxin Ma, Jiangyan Yi, Jianhua Tao, Ye~Bai, Zhengkun Tian, and Chenglong Wang.
\newblock Continual learning for fake audio detection.
\newblock {\em arXiv preprint arXiv:2104.07286}, 2021.

\bibitem{madaan2022representational}
Divyam Madaan, Jaehong Yoon, Yuanchun Li, Yunxin Liu, and Sung~Ju Hwang.
\newblock Representational continuity for unsupervised continual learning.
\newblock In {\em International Conference on Learning Representations (ICLR)}, 2022.

\bibitem{mi2020continual}
Fei Mi, Liangwei Chen, Mengjie Zhao, Minlie Huang, and Boi Faltings.
\newblock Continual learning for natural language generation in task-oriented dialog systems.
\newblock In {\em Findings of the Association for Computational Linguistics: EMNLP 2020}, volume {EMNLP} 2020, pages 3461--3474, 2020.

\bibitem{mo2023class}
Shentong Mo, Weiguo Pian, and Yapeng Tian.
\newblock Class-incremental grouping network for continual audio-visual learning.
\newblock In {\em Proceedings of the IEEE/CVF International Conference on Computer Vision}, pages 7788--7798, 2023.

\bibitem{mo2023audio}
Shentong Mo and Yapeng Tian.
\newblock Audio-visual grouping network for sound localization from mixtures.
\newblock In {\em Proceedings of the IEEE/CVF Conference on Computer Vision and Pattern Recognition}, pages 10565--10574, 2023.

\bibitem{nie2023bilateral}
Xing Nie, Shixiong Xu, Xiyan Liu, Gaofeng Meng, Chunlei Huo, and Shiming Xiang.
\newblock Bilateral memory consolidation for continual learning.
\newblock In {\em Proceedings of the IEEE/CVF Conference on Computer Vision and Pattern Recognition}, pages 16026--16035, 2023.

\bibitem{odena2017conditional}
Augustus Odena, Christopher Olah, and Jonathon Shlens.
\newblock Conditional image synthesis with auxiliary classifier gans.
\newblock In {\em International conference on machine learning}, pages 2642--2651. PMLR, 2017.

\bibitem{ostapenko2019learning}
Oleksiy Ostapenko, Mihai Puscas, Tassilo Klein, Patrick Jahnichen, and Moin Nabi.
\newblock Learning to remember: A synaptic plasticity driven framework for continual learning.
\newblock In {\em Proceedings of the IEEE/CVF conference on computer vision and pattern recognition}, pages 11321--11329, 2019.

\bibitem{owens2018audio}
Andrew Owens and Alexei~A Efros.
\newblock Audio-visual scene analysis with self-supervised multisensory features.
\newblock In {\em Proceedings of the European conference on computer vision (ECCV)}, pages 631--648, 2018.

\bibitem{park2021class}
Jaeyoo Park, Minsoo Kang, and Bohyung Han.
\newblock Class-incremental learning for action recognition in videos.
\newblock In {\em Proceedings of the IEEE/CVF International Conference on Computer Vision}, pages 13698--13707, 2021.

\bibitem{paszke2019pytorch}
Adam Paszke, Sam Gross, Francisco Massa, Adam Lerer, James Bradbury, Gregory Chanan, Trevor Killeen, Zeming Lin, Natalia Gimelshein, Luca Antiga, et~al.
\newblock Pytorch: An imperative style, high-performance deep learning library.
\newblock {\em Advances in neural information processing systems}, 32, 2019.

\bibitem{pham2021dualnet}
Quang Pham, Chenghao Liu, and Steven Hoi.
\newblock Dualnet: Continual learning, fast and slow.
\newblock {\em Advances in Neural Information Processing Systems}, 34:16131--16144, 2021.

\bibitem{pian2023audio}
Weiguo Pian, Shentong Mo, Yunhui Guo, and Yapeng Tian.
\newblock Audio-visual class-incremental learning.
\newblock In {\em Proceedings of the IEEE/CVF International Conference on Computer Vision}, pages 7799--7811, 2023.

\bibitem{prabhu2020gdumb}
Ameya Prabhu, Philip~HS Torr, and Puneet~K Dokania.
\newblock Gdumb: A simple approach that questions our progress in continual learning.
\newblock In {\em Computer Vision--ECCV 2020: 16th European Conference, Glasgow, UK, August 23--28, 2020, Proceedings, Part II 16}, pages 524--540. Springer, 2020.

\bibitem{purushwalkam2022challenges}
Senthil Purushwalkam, Pedro Morgado, and Abhinav Gupta.
\newblock The challenges of continuous self-supervised learning.
\newblock In {\em Computer Vision--ECCV 2022: 17th European Conference, Tel Aviv, Israel, October 23--27, 2022, Proceedings, Part XXVI}, pages 702--721. Springer, 2022.

\bibitem{radford2021learning}
Alec Radford, Jong~Wook Kim, Chris Hallacy, Aditya Ramesh, Gabriel Goh, Sandhini Agarwal, Girish Sastry, Amanda Askell, Pamela Mishkin, Jack Clark, et~al.
\newblock Learning transferable visual models from natural language supervision.
\newblock In {\em International conference on machine learning}, pages 8748--8763. PMLR, 2021.

\bibitem{rebuffi2017icarl}
Sylvestre-Alvise Rebuffi, Alexander Kolesnikov, Georg Sperl, and Christoph~H Lampert.
\newblock {iCaRL:} incremental classifier and representation learning.
\newblock In {\em Proceedings of the IEEE Conference on Computer Vision and Pattern Recognition}, pages 2001--2010, 2017.

\bibitem{ronneberger2015u}
Olaf Ronneberger, Philipp Fischer, and Thomas Brox.
\newblock U-net: Convolutional networks for biomedical image segmentation.
\newblock In {\em Medical Image Computing and Computer-Assisted Intervention--MICCAI 2015: 18th International Conference, Munich, Germany, October 5-9, 2015, Proceedings, Part III 18}, pages 234--241. Springer, 2015.

\bibitem{scialom2022fine}
Thomas Scialom, Tuhin Chakrabarty, and Smaranda Muresan.
\newblock Fine-tuned language models are continual learners.
\newblock In {\em Proceedings of the 2022 Conference on Empirical Methods in Natural Language Processing}, pages 6107--6122, 2022.

\bibitem{spence2007audiovisual}
Charles Spence.
\newblock Audiovisual multisensory integration.
\newblock {\em Acoustical science and technology}, 28(2):61--70, 2007.

\bibitem{srinivasan2022climb}
Tejas Srinivasan, Ting-Yun Chang, Leticia Leonor~Pinto Alva, Georgios Chochlakis, Mohammad Rostami, and Jesse Thomason.
\newblock {CL}i{MB}: A continual learning benchmark for vision-and-language tasks.
\newblock In {\em Thirty-sixth Conference on Neural Information Processing Systems Datasets and Benchmarks Track}, 2022.

\bibitem{stein2008multisensory}
Barry~E Stein and Terrence~R Stanford.
\newblock Multisensory integration: current issues from the perspective of the single neuron.
\newblock {\em Nature reviews neuroscience}, 9(4):255--266, 2008.

\bibitem{su2023separating}
Yiyang Su, Ali Vosoughi, Shijian Deng, Yapeng Tian, and Chenliang Xu.
\newblock Separating invisible sounds toward universal audiovisual scene-aware sound separation.
\newblock {\em arXiv preprint arXiv:2310.11713}, 2023.

\bibitem{sussman2005integration}
Elyse~S Sussman.
\newblock Integration and segregation in auditory scene analysis.
\newblock {\em The Journal of the Acoustical Society of America}, 117(3):1285--1298, 2005.

\bibitem{tan2023language}
Reuben Tan, Arijit Ray, Andrea Burns, Bryan~A Plummer, Justin Salamon, Oriol Nieto, Bryan Russell, and Kate Saenko.
\newblock Language-guided audio-visual source separation via trimodal consistency.
\newblock In {\em Proceedings of the IEEE/CVF Conference on Computer Vision and Pattern Recognition}, pages 10575--10584, 2023.

\bibitem{tang2022learning}
Yu-Ming Tang, Yi-Xing Peng, and Wei-Shi Zheng.
\newblock Learning to imagine: Diversify memory for incremental learning using unlabeled data.
\newblock In {\em Proceedings of the IEEE/CVF Conference on Computer Vision and Pattern Recognition}, pages 9549--9558, 2022.

\bibitem{tian2021cyclic}
Yapeng Tian, Di~Hu, and Chenliang Xu.
\newblock Cyclic co-learning of sounding object visual grounding and sound separation.
\newblock In {\em Proceedings of the IEEE/CVF Conference on Computer Vision and Pattern Recognition}, pages 2745--2754, 2021.

\bibitem{tian2018audio}
Yapeng Tian, Jing Shi, Bochen Li, Zhiyao Duan, and Chenliang Xu.
\newblock Audio-visual event localization in unconstrained videos.
\newblock In {\em Proceedings of the European conference on computer vision (ECCV)}, pages 247--263, 2018.

\bibitem{tong2022videomae}
Zhan Tong, Yibing Song, Jue Wang, and Limin Wang.
\newblock Videomae: Masked autoencoders are data-efficient learners for self-supervised video pre-training.
\newblock {\em Advances in neural information processing systems}, 35:10078--10093, 2022.

\bibitem{tzinis2020into}
Efthymios Tzinis, Scott Wisdom, Aren Jansen, Shawn Hershey, Tal Remez, Daniel~PW Ellis, and John~R Hershey.
\newblock Into the wild with audioscope: Unsupervised audio-visual separation of on-screen sounds.
\newblock {\em arXiv preprint arXiv:2011.01143}, 2020.

\bibitem{tzinis2022audioscopev2}
Efthymios Tzinis, Scott Wisdom, Tal Remez, and John~R Hershey.
\newblock Audioscopev2: Audio-visual attention architectures for calibrated open-domain on-screen sound separation.
\newblock In {\em European Conference on Computer Vision}, pages 368--385. Springer, 2022.

\bibitem{wang2022foster}
Fu{-}Yun Wang, Da{-}Wei Zhou, Han{-}Jia Ye, and De{-}Chuan Zhan.
\newblock {FOSTER:} feature boosting and compression for class-incremental learning.
\newblock In {\em In Proceedings of the European Conference on Computer Vision (ECCV)}, pages 398--414, 2022.

\bibitem{wang2021few}
Yu~Wang, Nicholas~J Bryan, Mark Cartwright, Juan~Pablo Bello, and Justin Salamon.
\newblock Few-shot continual learning for audio classification.
\newblock In {\em ICASSP 2021-2021 IEEE International Conference on Acoustics, Speech and Signal Processing (ICASSP)}, pages 321--325. IEEE, 2021.

\bibitem{wang2022learningrepresentations}
Zhepei Wang, Cem Subakan, Xilin Jiang, Junkai Wu, Efthymios Tzinis, Mirco Ravanelli, and Paris Smaragdis.
\newblock Learning representations for new sound classes with continual self-supervised learning.
\newblock {\em IEEE Signal Processing Letters}, 29:2607--2611, 2022.

\bibitem{wang2019continual}
Zhepei Wang, Cem Subakan, Efthymios Tzinis, Paris Smaragdis, and Laurent Charlin.
\newblock Continual learning of new sound classes using generative replay.
\newblock In {\em 2019 IEEE Workshop on Applications of Signal Processing to Audio and Acoustics (WASPAA)}, pages 308--312. IEEE, 2019.

\bibitem{wang2022learning}
Zifeng Wang, Zizhao Zhang, Chen-Yu Lee, Han Zhang, Ruoxi Sun, Xiaoqi Ren, Guolong Su, Vincent Perot, Jennifer Dy, and Tomas Pfister.
\newblock Learning to prompt for continual learning.
\newblock In {\em Proceedings of the IEEE/CVF Conference on Computer Vision and Pattern Recognition}, pages 139--149, 2022.

\bibitem{weisser2018complex}
Adam Weisser.
\newblock {\em Complex acoustic environments: Concepts, methods, and auditory perception}.
\newblock PhD thesis, PhD thesis). Macquarie University, Sydney, Australia. doi: 1959.14/1266534, 2018.

\bibitem{xiao2023endpoints}
Jia-Wen Xiao, Chang-Bin Zhang, Jiekang Feng, Xialei Liu, Joost van~de Weijer, and Ming-Ming Cheng.
\newblock Endpoints weight fusion for class incremental semantic segmentation.
\newblock In {\em Proceedings of the IEEE/CVF Conference on Computer Vision and Pattern Recognition}, pages 7204--7213, 2023.

\bibitem{xu2019recursive}
Xudong Xu, Bo~Dai, and Dahua Lin.
\newblock Recursive visual sound separation using minus-plus net.
\newblock In {\em Proceedings of the IEEE/CVF International Conference on Computer Vision}, pages 882--891, 2019.

\bibitem{yan2022generative}
Shipeng Yan, Lanqing Hong, Hang Xu, Jianhua Han, Tinne Tuytelaars, Zhenguo Li, and Xuming He.
\newblock Generative negative text replay for continual vision-language pretraining.
\newblock In {\em Computer Vision--ECCV 2022: 17th European Conference, Tel Aviv, Israel, October 23--27, 2022, Proceedings, Part XXXVI}, pages 22--38. Springer, 2022.

\bibitem{ye2023lavss}
Yuxin Ye, Wenming Yang, and Yapeng Tian.
\newblock Lavss: Location-guided audio-visual spatial audio separation.
\newblock {\em arXiv preprint arXiv:2310.20446}, 2023.

\bibitem{zhang2023you}
Xiaohui Zhang, Jiangyan Yi, Jianhua Tao, Chenglong Wang, and Chu~Yuan Zhang.
\newblock Do you remember? overcoming catastrophic forgetting for fake audio detection.
\newblock In {\em International Conference on Machine Learning}, pages 41819--41831. PMLR, 2023.

\bibitem{zhao2019sound}
Hang Zhao, Chuang Gan, Wei-Chiu Ma, and Antonio Torralba.
\newblock The sound of motions.
\newblock In {\em Proceedings of the IEEE International Conference on Computer Vision}, pages 1735--1744, 2019.

\bibitem{zhao2018sound}
Hang Zhao, Chuang Gan, Andrew Rouditchenko, Carl Vondrick, Josh McDermott, and Antonio Torralba.
\newblock The sound of pixels.
\newblock In {\em The European Conference on Computer Vision (ECCV)}, September 2018.

\bibitem{zhou2023deep}
Da-Wei Zhou, Qi-Wei Wang, Zhi-Hong Qi, Han-Jia Ye, De-Chuan Zhan, and Ziwei Liu.
\newblock Deep class-incremental learning: A survey.
\newblock {\em arXiv preprint arXiv:2302.03648}, 2023.

\bibitem{zhou2023learning}
Da-Wei Zhou, Yuanhan Zhang, Jingyi Ning, Han-Jia Ye, De-Chuan Zhan, and Ziwei Liu.
\newblock Learning without forgetting for vision-language models.
\newblock {\em arXiv preprint arXiv:2305.19270}, 2023.

\bibitem{zhou2022detecting}
Xingyi Zhou, Rohit Girdhar, Armand Joulin, Philipp Kr{\"a}henb{\"u}hl, and Ishan Misra.
\newblock Detecting twenty-thousand classes using image-level supervision.
\newblock In {\em European Conference on Computer Vision}, pages 350--368. Springer, 2022.

\bibitem{zhu2022visually}
Lingyu Zhu and Esa Rahtu.
\newblock Visually guided sound source separation and localization using self-supervised motion representations.
\newblock In {\em Proceedings of the IEEE/CVF Winter Conference on Applications of Computer Vision}, pages 1289--1299, 2022.

\end{thebibliography}





\clearpage

\appendix


\appendix

\section{Appendix}
In this appendix, we present a more detailed model architecture in section~\ref{sec:architecture}. Following that, we show the experimental comparison to the uni-modal baseline to prove the effectiveness of cross-modal similarity modeling and preservation in section~\ref{sec:unimodal}. After that, in section~\ref{sec:ave_vgg}, we conduct experiments on the AVE~\cite{tian2018audio} and the VGGSound~\cite{chen2020vggsound} datasets, which includes a broader range of audio-visual data beyond the music domain. Furthermore, in section~\ref{sec:old_class}, we present the performance on old classes of our proposed method and baselines, to further prove that our method can better mitigate the catastrophic forgetting problem compared to baselines. Finally, we offer additional visualization results in section~\ref{sec:vis}, to further demonstrate that our method can better handle the catastrophic forgetting problem in the proposed Continual Audio-Visual Sound Separation task compared to other continual learning baseline methods.

\subsection{Model Architecture}
\label{sec:architecture}
\paragraph{Audio Network.} Following~\cite{chen2023iquery}, we use the U-Net~\cite{ronneberger2015u} framework as the architecture of our audio network. In our experiments, the audio network has 7 down-convolutions and 7 up-convolutions. It takes a 2D Time-Frequency spectrum with size of $1\times 256\times 256$ as an input to generate the latent representation with a size of $256\times 32\times 32$ through the encoder part. Finally, it outputs the audio embedding with a size of $32\times 256\times 256$ through the decoder.

\paragraph{Audio-Visual Transformer.} We follow the implementation in~\cite{chen2023iquery}, and use the Transformer decoder architecture as our audio-visual Transformer. The audio-visual Transformer consists of 4 decoder layers, with the first layer being the motion cross-attention layer and the following 3 layers performing audio cross-attention and self-attention operation. The audio-visual Transformer generates the separated audio feature with a dimension of $256$, followed by a two-layers MLP to obtain the mask embedding with a dimension of $32$. Then, the channel-wise multiplication is applied between the generated mask embedding and the audio embedding to obtain the predicted mask $\hat{\boldsymbol{M}}$.

\paragraph{Video Encoder $\&$ Object Detector $\&$ Object Encoder.} For the pre-trained video encoder, object detector, and object encoder, we use the VideoMAE~\cite{tong2022videomae}, Detic~\cite{zhou2022detecting}, and CLIP~\cite{radford2021learning}, respectively. These models are frozen during the training process and the input size, out size, and internal feature dimensions remain the same as in their original implementations.

\subsection{Compared to Uni-modal Baseline}
\label{sec:unimodal}
To evaluate the superiority of cross-modal similarity modeling and preservation in continual audio-visual sound separation, in this subsection, we constructed a variant of our ContAV-Sep, in which we modify our proposed CrossSDC to the intra-modal similarity distillation version. We name it as ContAV-Sep-intra. The experimental results are shown in Tab.~\ref{tab:intra}. We can see that, our ContAV-Sep outperforms the variant significantly, further validating the effectiveness of modeling and preserving cross-modal similarity.

\begin{table}[htbp]
  \centering
  \caption{Comparison to the uni-modal variant on MUSIC-21 dataset.}
    \begin{tabular}{lccc}
    \toprule
    Methods & SDR$\uparrow$   & SIR$\uparrow$   & SAR$\uparrow$ \\
    \midrule
    ContAV-Sep-intra & 6.86 & 13.13  & 12.31 \\
    \textbf{ContAV-Sep} & \textbf{7.33}  & \textbf{13.55}  & \textbf{13.01} \\
    \bottomrule
    \end{tabular}%
  \label{tab:intra}%
\end{table}%


\subsection{Experiments on the AVE and the VGGSound datasets}
\label{sec:ave_vgg}
To further evaluate the efficacy of our proposed method across a broader sound domain, we conduct experiments using the AVE~\cite{tian2018audio} and the VGGSound~\cite{chen2020vggsound} datasets. In the experiments on the AVE dataset, we randomly split the 28 classes in the AVE dataset into 4 tasks, each of which contains 7 classes. The results are presented in Tab.~\ref{tab:AVE}, in which our ContAV-Sep outperforms the baseline models in terms of the SDR and SIR metrics, further demonstrating the robustness of our method beyond the domain of musical sounds. However, it was noted that both our method and the upper bound exhibit relatively low SAR scores when compared to the baselines. Gao et al.~\cite{gao2019co} provide an interpretation for this phenomenon, explaining that the SAR primarily captures the absence of artifacts. Therefore, it can remain high even when the separation quality is suboptimal. In contrast, the SDR and SIR metrics are used to evaluate the accuracy of the separation. For the experiments on the VGGSound~\cite{chen2020vggsound} dataset, we follow~\cite{pian2023audio} and randomly selecting 100 classes for continual learning. These classes are divided into 4 tasks, each containing 25 classes. Given the significantly larger number of samples per class in the VGGSound dataset, we set the memory size to 20 samples per class for methods that utilize memory. The experimental results, shown in Tab.~\ref{tab:VGG}, demonstrate that our proposed ContAV-Sep consistently outperforms the baseline methods on the VGGSound dataset in the context of continual audio-visual sound separation.

\begin{table}[htbp]
  \centering
  \caption{Continual audio-visual separation results on the AVE~\cite{tian2018audio} dataset.}
    \begin{tabular}{lccc}
    \toprule
    Methods & SDR$\uparrow$   & SIR$\uparrow$   & SAR$\uparrow$ \\
    \midrule
    iQuery~\cite{chen2023iquery} + Fine-tuning & 2.07  & 5.64 & \textbf{12.83} \\
    iQuery~\cite{chen2023iquery} + LwF (w/ memory)~\cite{li2017learning} & 2.19 & 6.43 & 10.67 \\
    iQuery~\cite{chen2023iquery} + PLOP (w/ memory)~\cite{douillard2021plop} & 2.45 & 6.11 & 11.57 \\
    iQuery~\cite{chen2023iquery} + AV-CIL (w/ memory)~\cite{pian2023audio} & 2.53 & 6.64 & 11.26 \\
    \textbf{ContAV-Sep (Ours)} & \textbf{2.72} & \textbf{7.32} & 9.86 \\
    \midrule
    Upper Bound (with iQuery) & 3.55  & 8.53 & 9.63 \\
    \bottomrule
    \end{tabular}%
  \label{tab:AVE}%
\end{table}%

\begin{table}[htbp]
  \centering
  \caption{Continual audio-visual separation results on the VGGSound~\cite{chen2020vggsound} dataset.}
    \begin{tabular}{lccc}
    \toprule
    Methods & SDR$\uparrow$   & SIR$\uparrow$   & SAR$\uparrow$ \\
    \midrule
    iQuery~\cite{chen2023iquery} + Fine-tuning & 3.69 & 7.23 & \textbf{12.84} \\
    iQuery~\cite{chen2023iquery} + LwF (w/ memory)~\cite{li2017learning} & 4.71 & 8.89 & 11.70 \\
    iQuery~\cite{chen2023iquery} + PLOP (w/ memory)~\cite{douillard2021plop} & 4.56 & 8.32 & 12.34 \\
    iQuery~\cite{chen2023iquery} + AV-CIL (w/ memory)~\cite{pian2023audio} & 4.69 & 8.61 & 11.60 \\
    \textbf{ContAV-Sep (Ours)} & \textbf{4.90} & \textbf{9.25} & 11.73 \\
    \midrule
    Upper Bound (with iQuery) & 5.57 & 9.19 & 13.24 \\
    \bottomrule
    \end{tabular}%
  \label{tab:VGG}%
\end{table}%

\subsection{Results on Old Classes}
\label{sec:old_class}
To further evaluate our proposed method's ability to handle the catastrophic forgetting problem, we use $SDR_t^{old}$, $SIR_t^{old}$, and $SAR_t^{old}$ to denote the separation performance on old classes after training at incremental step $t$. And then, we average each of these three metrics over all incremental steps, yielding $SDR_{mean}^{old}$, $SIR_{mean}^{old}$, and $SAR_{mean}^{old}$:
\begin{equation}
    \begin{split}
        SDR_{mean}^{old} = \frac{1}{t-1}\sum^t_{i=2} SDR_t^{old}, \\
        SIR_{mean}^{old} = \frac{1}{t-1}\sum^t_{i=2} SIR_t^{old}, \\
        SAR_{mean}^{old} = \frac{1}{t-1}\sum^t_{i=2} SAR_t^{old}.
    \end{split}
    \label{eq:metrics}
\end{equation}
The results are presented in Tab.~\ref{tab:old_class}, where it is evident that our method outperforms the baseline models, indicating a superior capability in addressing catastrophic forgetting within the context of continual audio-visual sound separation.

\begin{table}[htbp]
  \centering
  \caption{Experimental results on old classes on MUSIC-21 dataset.}
    \begin{tabular}{lccc}
    \toprule
    Methods & $SDR^{old}_{mean}\uparrow$   & $SIR^{old}_{mean}\uparrow$   & $SAR^{old}_{mean}\uparrow$ \\
    \midrule
    iQuery~\cite{chen2023iquery} + LwF~\cite{li2017learning} (w/ memory)  & 7.61 & 13.76 & 12.90 \\
    iQuery~\cite{chen2023iquery} + PLOP~\cite{douillard2021plop} (w/ memory) & 7.90 & 14.05 & 12.54 \\
    iQuery~\cite{chen2023iquery} + EWF~\cite{xiao2023endpoints} (w/ memory) & 6.96 & 13.20 & 12.86 \\
    \textbf{ContAV-Sep (Ours)} & \textbf{8.11} & \textbf{14.30} & \textbf{13.26} \\
    \bottomrule
    \end{tabular}%
  \label{tab:old_class}%
\end{table}%

\subsection{Visualization Results}
\label{sec:vis}

\begin{figure*}
    \centering
    \includegraphics[width=0.89\textwidth]{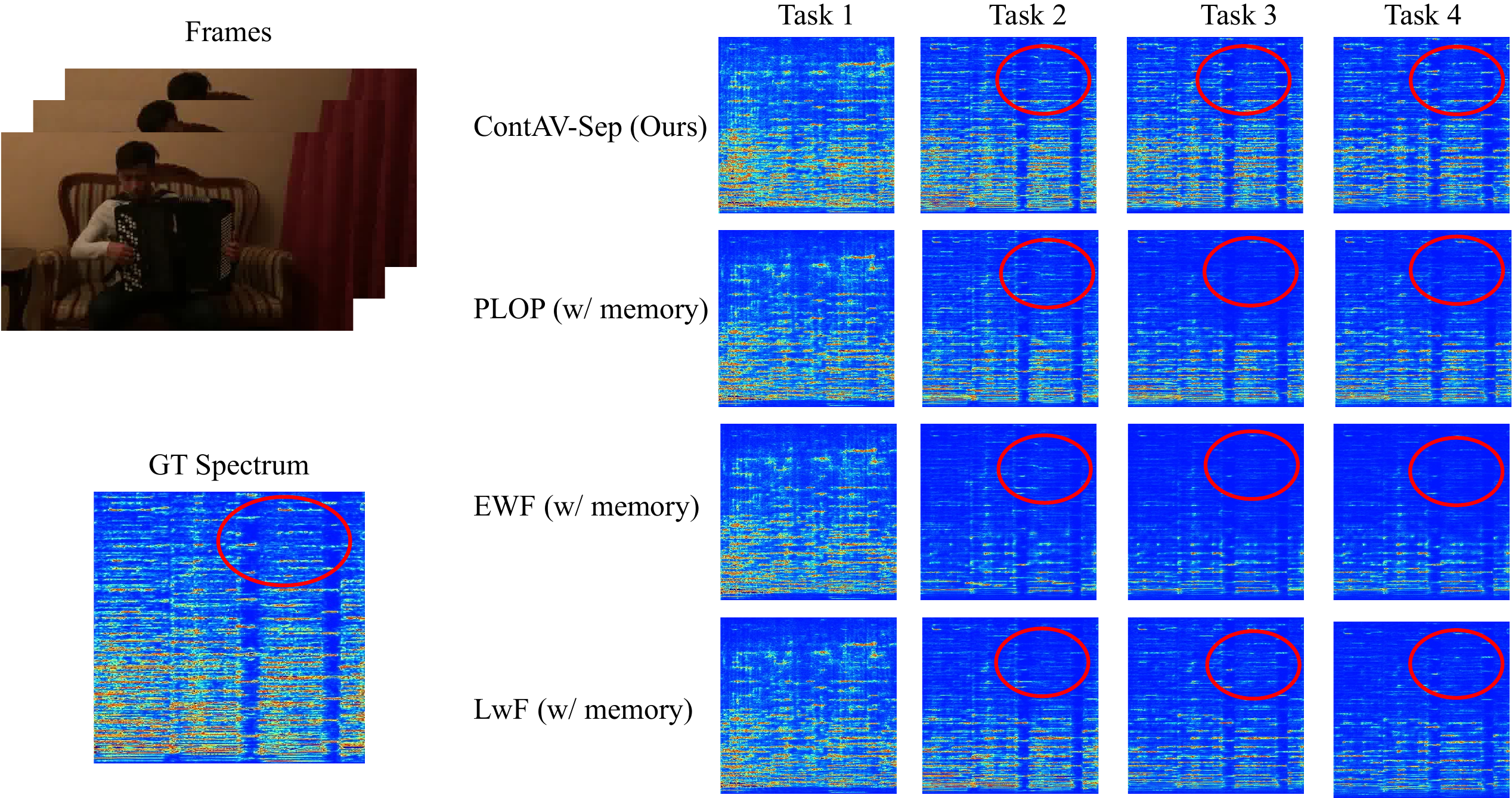}
    \caption{Left: a randomly selected sample with its frame and ground-truth spectrogram. Right: separated sounds by our ContAV-Sep and baselines at each incremental step.}
    \label{fig:vis}
\end{figure*}

\begin{figure*}
    \centering
    \includegraphics[width=0.89\textwidth]{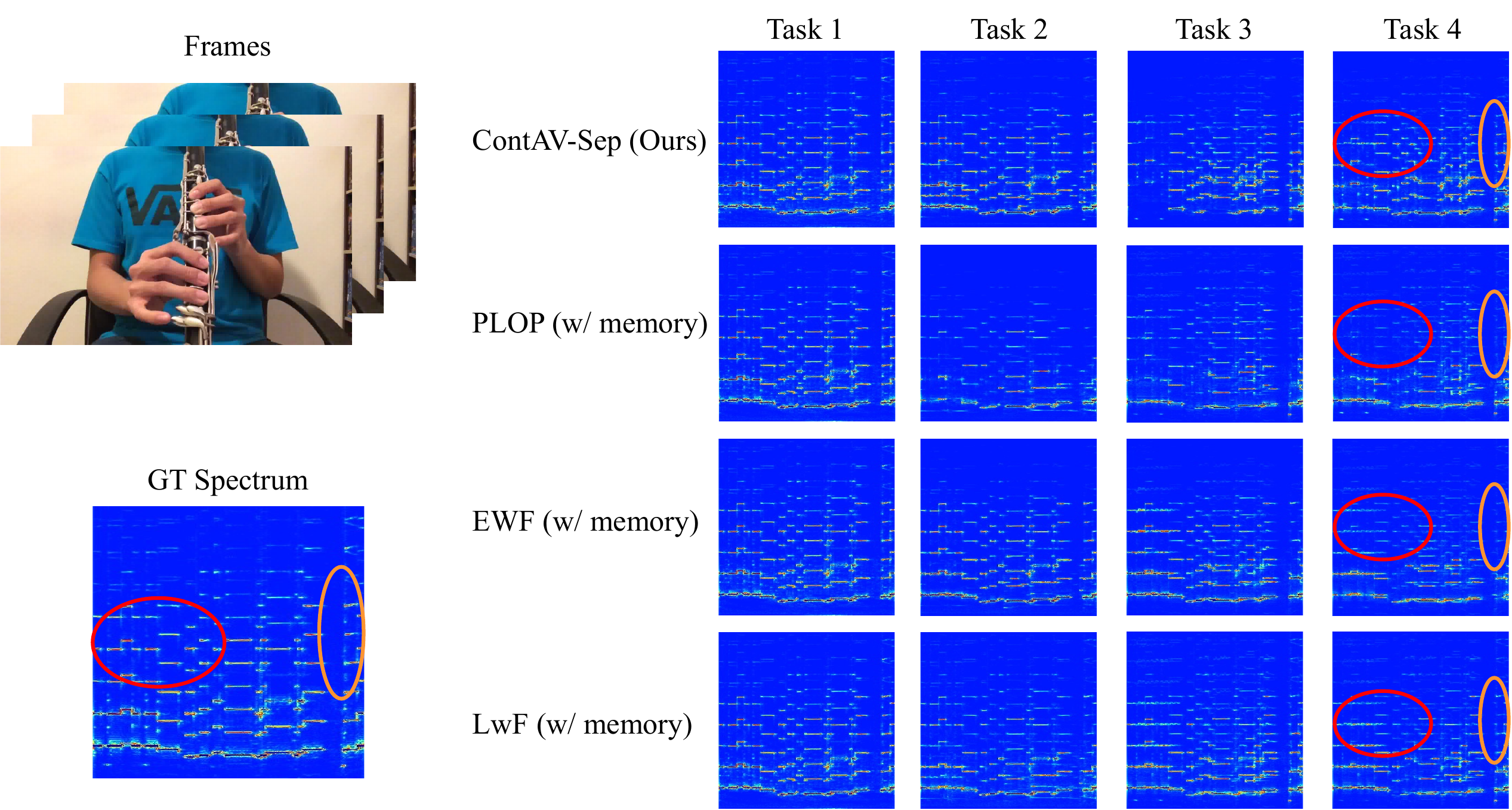}
    \caption{Left: A randomly selected sample with its frame and ground-truth spectrum. Right: Visualization of the separated sound by our ContAV-Sep and baselines at each incremental step.}
    \label{fig:vis_1}
\end{figure*}

\begin{figure*}
    \centering
    \includegraphics[width=0.89\textwidth]{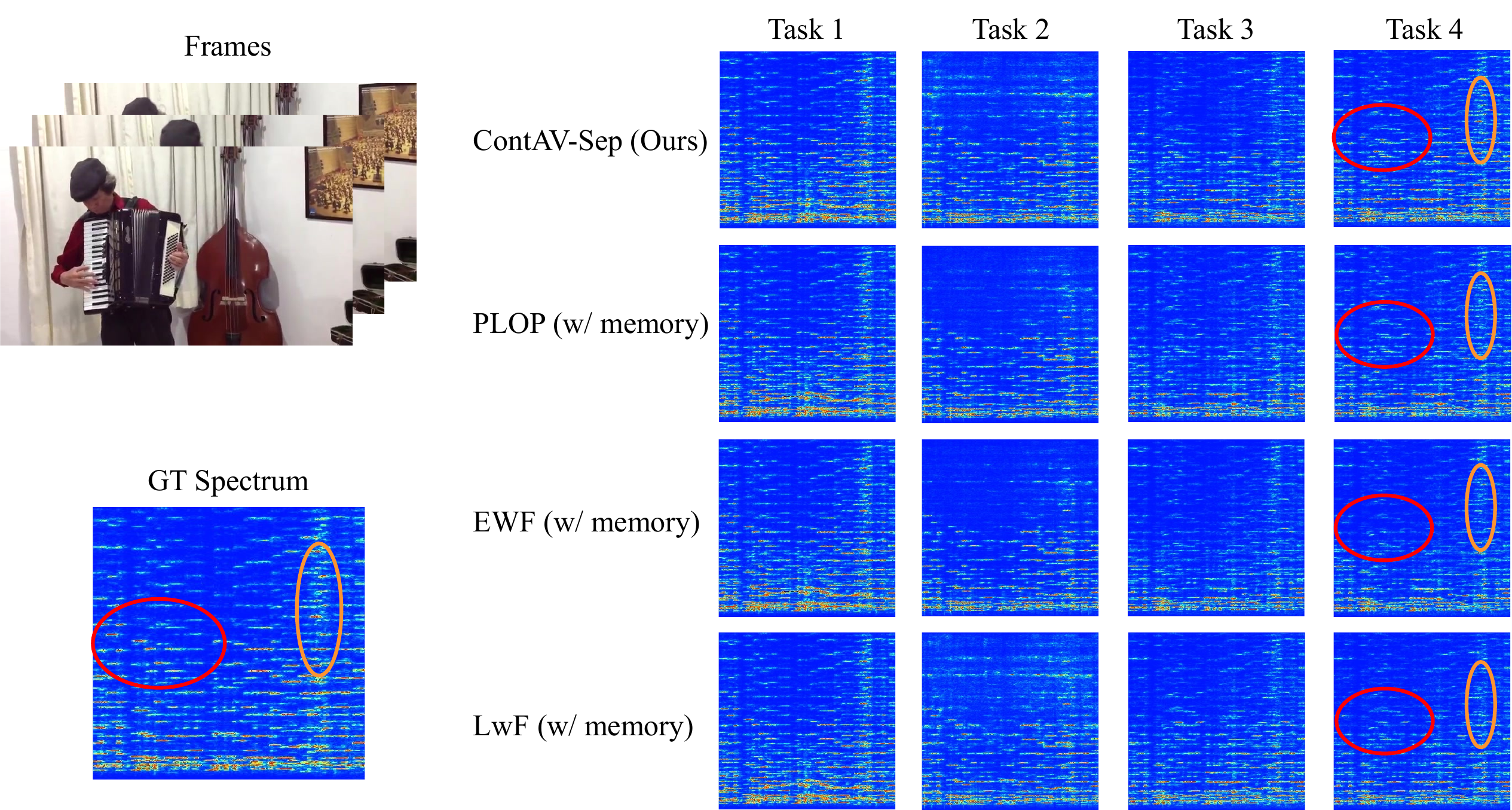}
    \caption{Left: A randomly selected sample with its frame and ground-truth spectrum. Right: Visualization of the separated sound by our ContAV-Sep and baselines at each incremental step.}
    \label{fig:vis_2}
\end{figure*}

\begin{figure*}
    \centering
    \includegraphics[width=0.89\textwidth]{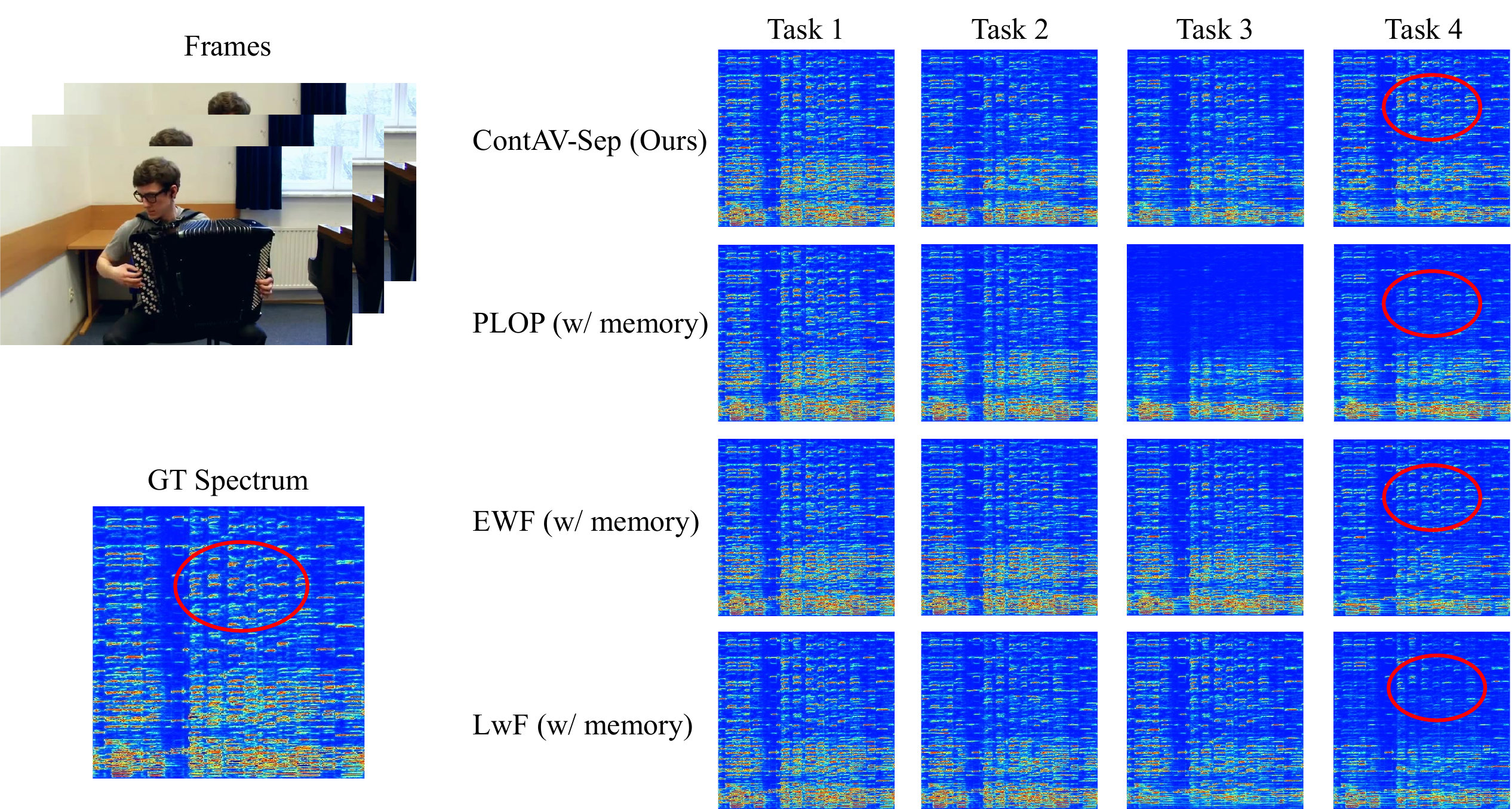}
    \caption{Left: A randomly selected sample with its frame and ground-truth spectrum. Right: Visualization of the separated sound by our ContAV-Sep and baselines at each incremental step.}
    \label{fig:vis_3}
\end{figure*}

\begin{figure*}
    \centering
    \includegraphics[width=0.89\textwidth]{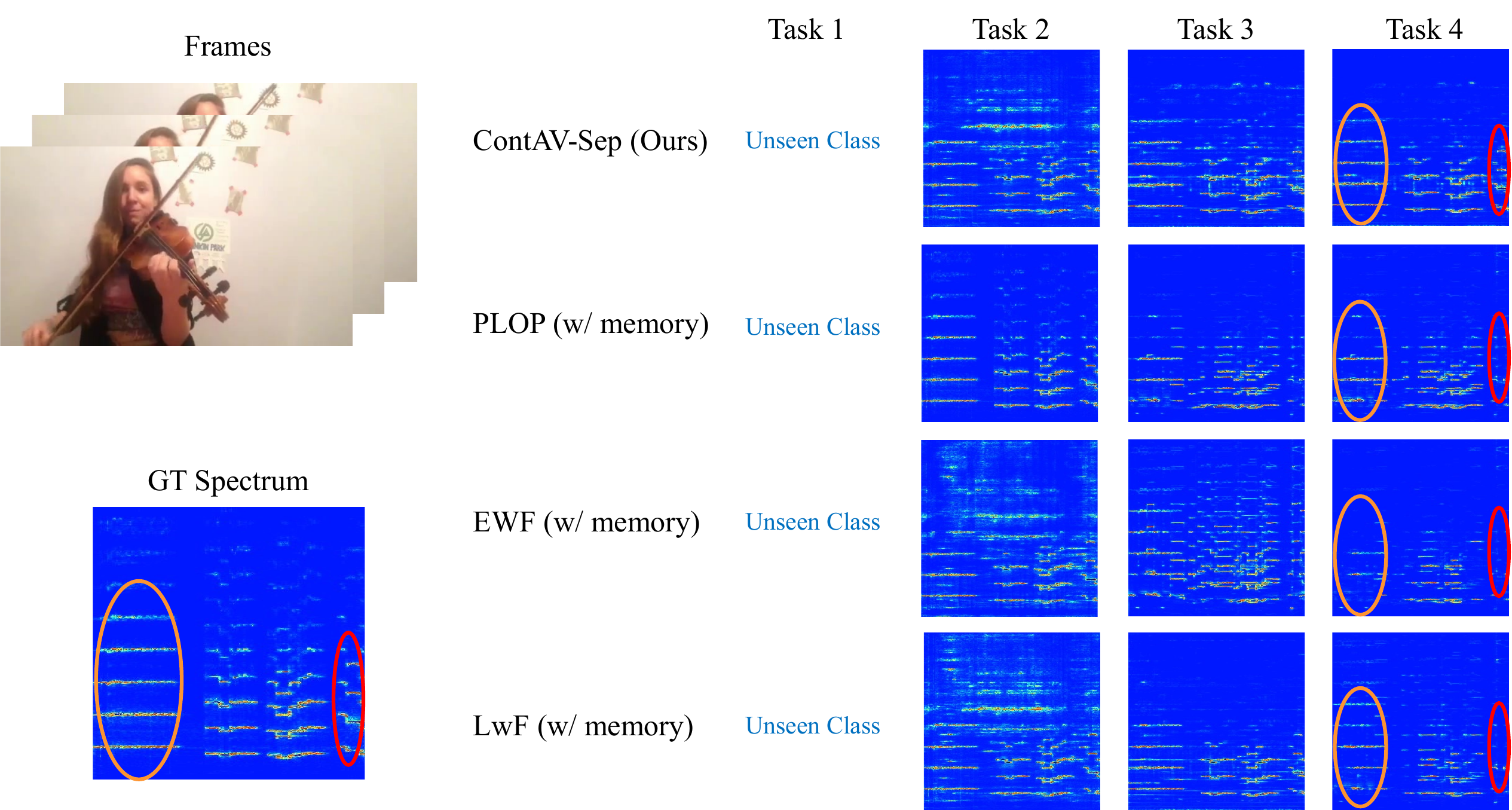}
    \caption{Left: A randomly selected sample with its frame and ground-truth spectrum. Right: Visualization of the separated sound by our ContAV-Sep and baselines at each incremental step.}
    \label{fig:vis_4}
\end{figure*}

\begin{figure*}
    \centering
    \includegraphics[width=0.89\textwidth]{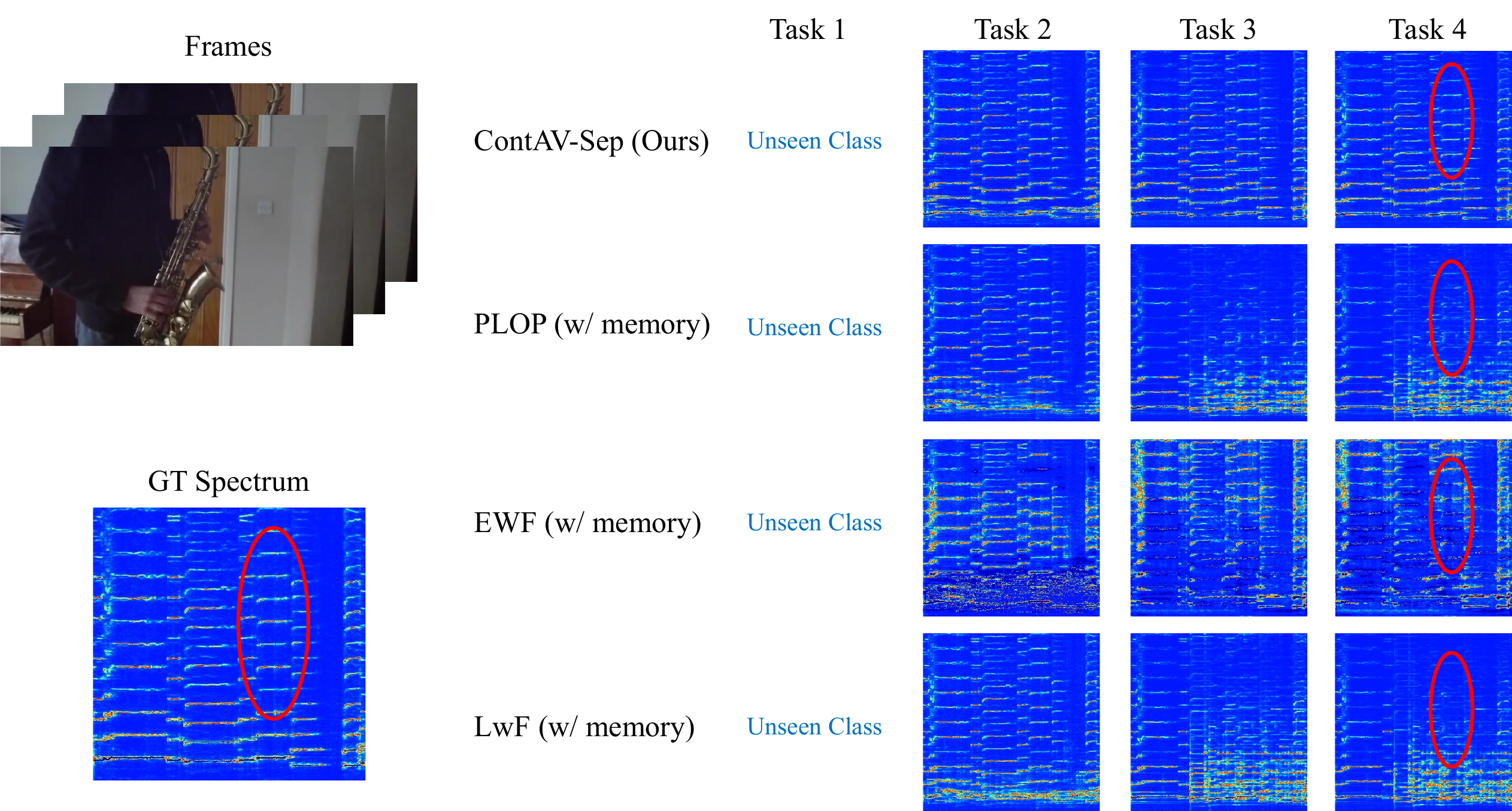}
    \caption{Left: A randomly selected sample with its frame and ground-truth spectrum. Right: Visualization of the separated sound by our ContAV-Sep and baselines at each incremental step.}
    \label{fig:vis_5}
\end{figure*}

We present visualization results in Fig.~\ref{fig:vis},~\ref{fig:vis_1},~\ref{fig:vis_2},~\ref{fig:vis_3},~\ref{fig:vis_4}, and~\ref{fig:vis_5}. In each figure, we show the results of our proposed ContAV-Sep, along with four baseline methods PLOP~\cite{douillard2021plop}, EWF~\cite{xiao2023endpoints}, and Lwf~\cite{li2017learning} at each incremental step. We highlight specific areas after the final step, where it is evident that our method yields a more accurate predicted spectrum and preserves more details after training on new tasks, which demonstrates that our method can better handle the catastrophic forgetting problem compared to baseline continual learning methods.




\end{document}